\ifdefined\pdfoutput\pdfoutput=1\fi
\documentclass{article}
\usepackage{arxiv}
\usepackage[utf8]{inputenc}
\usepackage[T1]{fontenc}
\usepackage{amsmath,amssymb,booktabs,graphicx,float,placeins,url}
\usepackage{natbib}
\usepackage{hyperref}
\usepackage{microtype}
\hypersetup{hidelinks,
  pdftitle={2D-FET-Bench: From Spatial Reasoning to FET Design on Flakes},
  pdfauthor={Dunzhi Zhou, Chengyu Zhu, Gang Qiu, Caiwen Ding}}
\newcommand{\benchmark}{\textsc{2D-FET-Bench}}
\title{2D-FET-Bench: From Spatial Reasoning to FET Design on Flakes}
\renewcommand{\shorttitle}{2D-FET-Bench}
\renewcommand{\undertitle}{Preprint}
\renewcommand{\headeright}{Preprint}
\author{Dunzhi Zhou$^{1}$ \quad Chengyu Zhu$^{1}$ \quad Gang Qiu$^{1}$ \quad Caiwen Ding$^{1,2}$ \\
$^{1}$University of Minnesota, Minneapolis, MN, USA \\
$^{2}$Future Intelligence Labs \\
\texttt{\{zhou2299, zhu00952, gqiu, dingc\}@umn.edu}}
\date{}
\begin{document}
\maketitle

\begin{abstract}
Field-effect transistor (FET) layouts on exfoliated two-dimensional flakes are
typically drawn by hand for each flake, placing contacts and gates to match its
position and outline in optical micrographs. To our knowledge, no executable benchmark
tests whether language-model agents can perform this flake-specific
construction reliably. We introduce 2D-FET-Bench V2, a benchmark of 128 layout
tasks built from microscopy-derived flake contours, including hole-containing
flakes and multi-flake tasks. Each task supplies a textual device specification
and contour coordinates. An agent generates typed polygon and path operations
rendered to GDSII. A deterministic verifier checks geometric and structural
requirements, and a separate integrity check verifies that the supplied
contours remain unchanged. Scripted reference layouts pass all 128 tasks,
showing that every task is solvable. We evaluate six models and seven
workflow and scaffold variants of GPT5.6-Luna, with five attempts per task. The
best-performing configuration in the six-model panel, GPT5.6-Luna with ReAct-3,
passes 62.3\% of attempts and solves 80.5\% of tasks at least once (coverage)
and 43.8\% in all five attempts (consistency). ReAct-3 exceeds
the one-pass Plan-and-Execute by 27.0 pass@1 points at $2.46\times$ the
tokens. An expert audit of one sampled verifier-passing layout per covered
task, across five ReAct-3 configurations, accepts 56.4\%--63.5\% of them. The
benchmark evaluates geometric and structural FET layout construction.
\end{abstract}

\section{Introduction}
Preparing field-effect transistor (FET) layouts on exfoliated two-dimensional
(2D) materials such as single-layer MoS$_2$ \citep{radisavljevic2011mos2}
requires adapting contacts, gates, and other features to each flake's unique
geometry. Mechanical exfoliation yields flakes with highly irregular boundaries
and sizes, and transfer and placement add variation in position and
orientation, so each flake needs its own layout. Automated microscopy can
locate candidate flakes \citep{masubuchi2018automated,uslu2025maskterial}, and a
specialized system can generate layouts for fixed device recipes
\citep{lee2026mos2layout}. General-purpose language-model agents can instead
translate flexible textual specifications and contour coordinates into
executable layout operations. Evaluating them requires repeated tests with
explicit geometric inputs and programmatically checkable success criteria.
\benchmark{} V2 provides such an evaluation. Each task supplies one or more
fixed flake contours as ordered coordinates, a textual device objective,
editable layer definitions, and layout rules. The agent returns typed polygon
and path operations that are interpreted into GDSII geometry. An attempt
succeeds only when the saved layout passes all geometric and structural checks
and preserves the supplied contours. Agents receive contour coordinates instead of
microscopy images, so the benchmark tests spatial reasoning from
specification to geometry separately from optical image perception.

V2 keeps the specification and interface structure, scaffolds, and workflows of
an earlier synthetic-polygon version (V1). It replaces the geometry with
microscopy-derived contours and adds hole probes
(Appendix~\ref{app:version-design}). We make three contributions.
\textbf{(1) An executable geometry-adaptive layout benchmark.} We introduce 128
tasks built from microscopy-derived graphene contours, including single- and
multi-region scenes with 1 to 10 flakes and 1 to 24 required transistors,
hole-containing flakes, multiple device structures, and
circuit-level templates. Construction combines solidity-stratified contour
selection with synthetic clustered placement for multi-region scenes
(Section~\ref{sec:construction}). \textbf{(2) Deterministic geometric and
structural verification.} Agents produce executable layout operations. A
KLayout-based verifier checks geometric rules and encoded structural
requirements, and a separate integrity test checks that the supplied contours
are unchanged. Scripted reference constructions pass the verifier for all 128
tasks, and an isolated re-score reproduces all 7,680 recorded attempt verdicts
(Section~\ref{sec:tasks}). \textbf{(3) Reliability across models, workflows, and
scaffolds.} We evaluate six models under ReAct-3/C0 and compare five C0
workflows and three ReAct-3 scaffolds within GPT5.6-Luna, using five attempts
per task. We report mean attempt success, task coverage, and observed
consistency, relate them to transistor demand, region count, flake geometry, and
device architecture, and add an expert audit of randomly sampled verifier-passing layouts
(Section~\ref{sec:results}).

\section{Related work}
Several benchmarks and agent systems address circuit and layout design.
GLayout provides an open evaluation set for language-driven analog layout with
design-rule and layout-versus-schematic results \citep{hammoud2024glayout}.
LayoutCopilot applies collaborating language-model agents to interactive analog
layout design \citep{liu2024layoutcopilot}. Rule2DRC tests agents that generate
DRC scripts, scored by execution on labeled GDSII layouts
\citep{kim2026rule2drc}. AnalogCoder is a training-free agent that writes
PySpice circuit code, evaluated on 24 analog design tasks
\citep{lai2025analogcoder}, and AMSbench covers circuit perception, analysis,
and design \citep{shi2025amsbench}. VerilogEval and RTLLM benchmark digital RTL
generation \citep{liu2023verilogeval,lu2024rtllm}. These benchmarks target
circuits or layouts on regular process geometry. Geometry-generation benchmarks
score CAD solids or diagrams. Text2CAD pairs text with parametric CAD sequences
\citep{khan2024text2cad}, and Text2CAD-Bench tests LLM-generated CAD across
geometric complexity \citep{wang2026text2cadbench}. BenchCAD assesses
executable CadQuery programs for industrial parts \citep{zhang2026benchcad}, and
CADTestBench runs tests of geometric and topological requirements on generated
CAD models \citep{mallis2026cadtests}. GGBench assesses textual construction
plans, code executability, and the geometric fidelity of drawn diagrams
\citep{wei2026ggbench}. In our benchmark, agents generate polygon and path
operations that produce layered GDSII geometry around fixed, irregular flake
contours, and deterministic geometric checks and a fixed-input integrity check
score the result. For evaluation, we follow the pass@k convention for repeated
code samples \citep{chen2021evaluating}. We report observed pass@5 for task
coverage and all-five for consistency across the same five attempts. We also
report resource use alongside success, as recommended for agent evaluation
\citep{kapoor2024agentsmatter}.

\section{Benchmark tasks and success criteria}
\label{sec:tasks}
\begin{figure}[t]
\centering\includegraphics[width=.92\textwidth,trim=0 17pt 0.10in 4pt,clip]{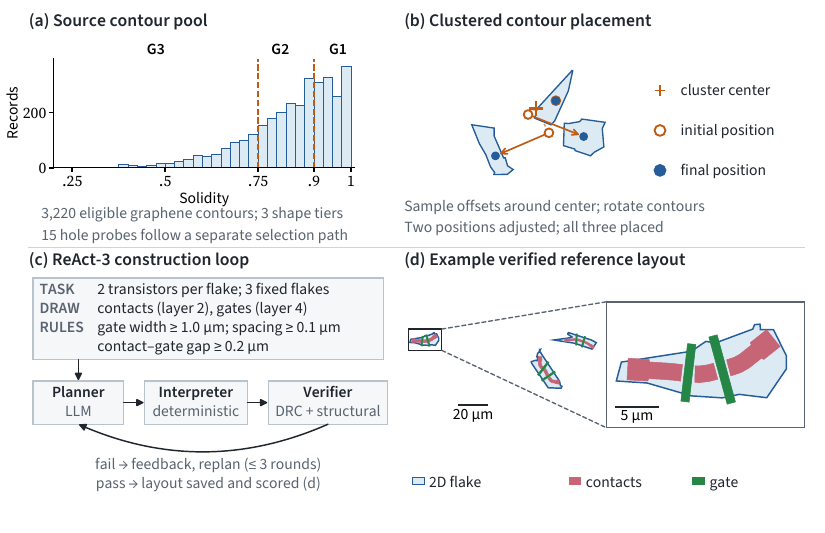}
\caption{Benchmark construction and task interface, illustrated with task
L3/case\_8. (a) The contour pool grouped by shape tier, with hole probes
selected separately. (b) An illustrative multi-flake placement that differs from
the placement of this task. (c) The ReAct-3 construction loop, with the
condensed request shown to the agent as its input. (d) The stored reference
layout for this task, produced by reference code. Contours derive from MaskTerial dataset
masks \citep{maskterial_dataset}, scales are nominal, and material labels assign the n- and p-type
device roles in each layout.}
\label{fig:interface}
\end{figure}

\paragraph{What the agent receives and constructs.}
Each task $t$ is a JSON file with two parts: an agent-visible specification
$x_t$ and scorer criteria $h_t$ withheld from the prompt. The
specification defines one or more fixed material regions (flakes) as polygons
in $\mu$m, including interior hole rings where present, and assigns each a
channel material (graphene, MoSe$_2$, or WSe$_2$). It also gives a textual
objective, a layer table, the design order in which layers are drawn, and
layout rules. Each layer-table entry gives a GDS layer and datatype number, a
name, a role (channel material, contact, gate electrode, via, pad cover, or
dielectric), and a status. Flake layers are read-only, and all other layers are
editable. The design order marks where the flakes sit in the stack (for
example, \texttt{[back\_gate, @graphene, contacts]}), so together with the
objective it fixes the device architecture: top-gated, local or global
back-gated, dual-gated, or edge-contacted. The task JSON tags each region's material,
but the prompt lists regions only by name and contour, so the agent reads
materials from the read-only layers. The agent's first prompt contains the objective (for example,
``Create 2 transistors per flake across 3 flakes''), the ordered contour
vertices to two decimal places, the layer table and design order, the layout
rules, and harness-derived constraints that suggest one gate-crossing location
and source/drain contact zones per flake. The prompt uses only text and
coordinates, and the scorer criteria $h_t$ stay with the evaluator. Each task is
attempted five times, indexed by $i\in\{1,\dots,5\}$. In attempt $i$, the agent
responds with typed polygon/path additions and cuts on the
editable layers, forming an operation sequence $a_{t,i}$. An interpreter
$\Phi_q$ applies $a_{t,i}$ to the fixed input geometry of $x_t$, producing
layered layout geometry $\mathcal G_{t,i}=\Phi_q(x_t,a_{t,i})$, saved as GDSII on
the $q=0.001\,\mu$m KLayout grid \citep{klayout}. Figure~\ref{fig:interface}(c,d)
shows one task. Its objective asks for two transistors on each of three
supplied flakes. The contours and material layers are fixed, and contacts
(layer 2) and gates (layer 4) are editable. The layout rules set a
$1.0\,\mu$m minimum gate width, $0.1\,\mu$m contact spacing, and $0.2\,\mu$m
contact--gate separation. The scorer checks the required gate crossings,
overlaps, and at least two transistors on each of the three flakes. The pictured layout was
built by the reference code for this task and passes all scorer checks with its
three input contours preserved.

\paragraph{Evaluation criteria and metrics.}
Attempt $i$ on task $t$ succeeds only if its saved layout $\mathcal G_{t,i}$
passes all design-rule checks (DRC) and structural layout checks and preserves
the fixed input geometry:
\begin{equation}
 s_{t,i}=\mathbf 1\!\left[\text{DRC}_{t,i}\land\text{Structural}_{t,i}\land
 \text{InputsPreserved}_{t,i}\right].
 \label{eq:success}
\end{equation}
Here $s_{t,i}\in\{0,1\}$ is the outcome of that attempt and $\mathbf 1[\cdot]$
is 1 when all three conditions hold. $\text{DRC}_{t,i}$ holds if
$\mathcal G_{t,i}$ satisfies the layout rules shown to the agent,
$\text{Structural}_{t,i}$ holds if it satisfies $h_t$ and contains nonempty
geometry on every editable layer in the design order, and
$\text{InputsPreserved}_{t,i}$ holds if its read-only layers still match the
flake polygons in $x_t$. A transistor is counted where a gate shape crosses a
flake and contact shapes touch the flake on both sides of the crossing. The
checks come from three sources (Table~\ref{tab:predicates}). Design-rule checks
such as minimum width, spacing, separation, and enclosure are listed as layout
rules in the prompt. Tasks with a via layer require contacts to enclose each via
by a margin, and the harness adds via--gate isolation rules that are also shown
to the agent. Structural checks on required overlaps, gate crossings,
per-flake transistor counts, and circuit-layout templates are encoded in the
scorer criteria $h_t$. Their content can appear in the objective or system
prompt, in harness guidance, or in verifier messages during revision.
Required-layer presence covers every editable layer in the design order (listed
as completeness in Table~\ref{tab:predicates}). Flake layers must stay
unchanged. After a GDS write/read round trip, the saved layout's fixed layers
are compared with the task initialization by XOR, without redrawing or
repairing them. This check runs on the saved file, separately from the verifier
used during the run, because the planner's operations are applied directly and
bypass the guarded drawing tools, which block edits to read-only layers. Across
all 7,680 recorded attempts, 7,642 (99.5\%) pass this final check. For task $t$,
let $c_t=\sum_{i=1}^{5}s_{t,i}$ be the number of successful attempts. Across $N$
tasks ($N=128$ unless stated otherwise), we report three task-averaged metrics
for each model configuration:
\begin{equation}
 \mathrm{pass@1}=\frac1N\sum_{t=1}^{N}\frac{c_t}{5},\qquad
 \mathrm{pass@5}=\frac1N\sum_{t=1}^{N}\mathbf 1[c_t\ge1],\qquad
 \mathrm{all\mbox{-}five}=\frac1N\sum_{t=1}^{N}\mathbf 1[c_t=5].
 \label{eq:passk}
\end{equation}
Pass@1 measures average attempt success, pass@5 measures task coverage within
five attempts, and all-five measures observed consistency across the same five
attempts of each task. An isolated re-score reproduces all 7,680 recorded
attempt verdicts, and all 128 reference layouts pass with their input contours
preserved. Benchmark success means that a layout meets the implemented
geometric and structural requirements and preserves its input contours.
Section~\ref{sec:expert-review} reports an expert audit of the intended device
structure in the sampled passing layouts, and Appendix~\ref{app:verifier} gives the
verifier's numerical behavior and the saved-layout preservation audit of all
7,680 recorded attempts.

\begin{table}[t]
\caption{Verification checks across 128 tasks. Check families are
design-rule checks (DRC), structural layout checks, and required-layer
completeness. Tasks is the number of tasks with the check, and total rules
counts its occurrences, excluding harness-added via--gate isolation rules.}
\label{tab:predicates}\centering\small
\begin{tabular}{llrrr}
\toprule
Verification check & Check family & Tasks & Total rules & \shortstack[r]{Median rules\\per task (when present)} \\
\midrule
\texttt{min\_width} & DRC & 127 & 132 & 1 \\
\texttt{min\_spacing} & DRC & 128 & 128 & 1 \\
\texttt{min\_separation} & DRC & 127 & 132 & 1 \\
\texttt{min\_enclosure} & DRC & 69 & 69 & 1 \\
\texttt{must\_not\_overlap} & DRC & 127 & 132 & 1 \\
\texttt{must\_touch\_boundary} & DRC & 5 & 8 & 1 \\
\addlinespace[2pt]
\texttt{must\_overlap} & Structural & 128 & 368 & 2 \\
\texttt{must\_cross} & Structural & 127 & 189 & 1 \\
\texttt{transistor\_count} & Structural & 127 & 369 & 2 \\
\texttt{circuit\_spec} & Structural & 26 & 26 & 1 \\
\addlinespace[2pt]
\texttt{layer\_not\_empty} & Completeness & 128 & 363 & 3 \\
\bottomrule
\end{tabular}

\end{table}

\section{Benchmark construction and statistics}
\label{sec:construction}
We construct the 128 tasks from graphene flakes in the semantic masks released
with the MaskTerial datasets \citep{maskterial_dataset,uslu2025maskterial}, one of
which reuses the graphene images of the 2DMatGMM datasets \citep{uslu2023twodmatgmm}.
Each flake is one connected mask component. Components touching the image border
or smaller than 300 pixels are discarded, and the remaining outlines are
smoothed into polygons. The main pool contains 3,220 hole-free contours of at
most 800\,$\mu$m$^2$, grouped into three shape tiers by solidity (outer-contour
area over convex-hull area): G1 ($\geq0.90$, near-convex), G2 (0.75--0.90,
moderately concave), and G3 (\mbox{$<0.75$}, strongly concave). Each task is
assigned a tier by a fixed cycle of roughly 30/40/30 within its template level
(non-top-gated templates use G1), and its contours are sampled at random
from that tier, keeping flakes that fit the required transistors. Specifications fix the number of flakes: 49 tasks contain one flake and 79
contain several. For multi-flake tasks, flakes are placed with a
Thomas-process clustering model fitted to flake centroids in
MaskTerial images, then screened for separation and
construction feasibility (Appendix~\ref{app:construction}).

The 128 tasks place 370 flakes
drawn from 352 flake IDs (347 distinct stored contours). The
median task contains two flakes and requires three transistors, and the tasks
span 1 to 10 flakes and 1 to 24 required transistors. Flakes have a median area
of 150.6\,$\mu$m$^2$ (26.5--796.7), a median solidity of 0.86 (0.31--1.00), and a
median of 13 contour vertices (4--51), and tasks have a median of 14
verification rules (3--26) and three required design layers (1--5). The
rendered task text (objective, layers, contours, and rules) has a median of 616 tokens (224--1,990). All source contours are
graphene-derived, and MoSe$_2$ and WSe$_2$ labels assign n-type and p-type
device roles to these contours. The 26 circuit tasks
also require circuit templates such as inverter and NAND/NOR structures, checked by
exact transistor counts and, for supported templates, contact sharing. For
NOR-type templates, the check also accepts a series arrangement. Other
transistor-count criteria set lower bounds per flake.

\begin{figure}[t]
\centering\includegraphics[width=\textwidth,trim=0 3pt 0 4pt,clip]{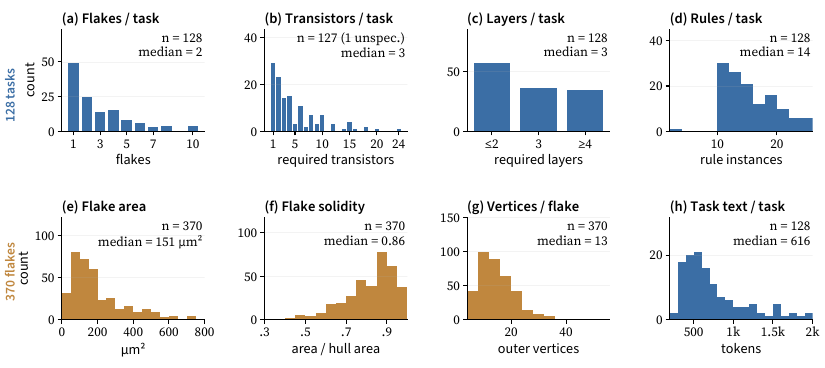}
\caption{Distributions over the 128 tasks (a: flakes, b: transistors, c: layers,
d: rules, h: task-text tokens) and their 370 flakes (e: area, f: solidity, g:
vertices). Each panel gives $n$ and the median. Flake areas use the nominal
$0.293\,\mu$m/pixel scale.}
\label{fig:statistics}
\end{figure}

Task specifications come
from V1: 97 keep their V1 structure with new geometry and MoS$_2$ renamed to
MoSe$_2$, 16 are restructured, and the 15 hole probes, which use holed contours
from a separate pool, are new. Each task also carries descriptive labels:
template level (L1--L4), difficulty class (D1--D4), and shape tier
(G1--G3). Template level identifies the task family. The difficulty class
combines task properties including required transistor count and was revised
after model results were available. These labels describe the constructed task
set. Figure~\ref{fig:interface}(a,b) summarizes construction,
Table~\ref{tab:composition} gives the six provenance groups, and
Figure~\ref{fig:statistics} and Table~\ref{tab:crosstab} characterize the
resulting tasks. Appendix~\ref{app:construction} documents sampling, scene
modifications, and scale assumptions, and Appendix~\ref{app:task-gallery} shows
the input flake contours of all 128 tasks, grouped by template level and drawn
with scale bars.

\begin{table}[htbp]
\caption{V2 task composition (128 tasks): counts, shares, and task-level
medians. Medians are over the tasks in each group. Required transistors
sum per-flake counts, and the task without a transistor-count criterion is
excluded from these medians. Group definitions are in
Appendix~\ref{app:composition-groups}.}
\label{tab:composition}\centering\small
\begin{tabular}{lrrr}
\toprule
Group & Tasks, n (\%) & \shortstack[r]{Median flakes\\per task} & \shortstack[r]{Median required\\transistors per task} \\
\midrule
Top-gated base & 55 (43.0\%) & 2.0 & 3.0 \\
Hole (15) and architecture (4) probes & 19 (14.8\%) & 1.0 & 1.0 \\
Gate/contact variants & 18 (14.1\%) & 2.0 & 2.0 \\
Repeated devices and circuits & 18 (14.1\%) & 3.0 & 9.0 \\
Three material roles & 12 (9.4\%) & 5.0 & 6.0 \\
Scene/constraint variants & 6 (4.7\%) & 2.0 & 2.5 \\
Overall & 128 (100.0\%) & 2.0 & 3.0 \\
\bottomrule
\end{tabular}

\end{table}

\begin{table}[t]
\caption{Composition of the 128 tasks by provenance group and by template level
(L), difficulty class (D), shape tier (G), and hole probes. Counts are tasks,
and rows sum to $n$ within each block.}
\label{tab:crosstab}\centering\small
\begingroup\setlength{\tabcolsep}{4pt}\setlength{\abovetopsep}{4pt}
\begin{tabular}{l r rrrr rrrr rrr r}
\toprule
Group & $n$ & \multicolumn{4}{c}{Level} & \multicolumn{4}{c}{Difficulty class} & \multicolumn{3}{c}{Shape tier} & Holes \\
\cmidrule(lr){3-6}\cmidrule(lr){7-10}\cmidrule(lr){11-13}
 & & L1 & L2 & L3 & L4 & D1 & D2 & D3 & D4 & G1 & G2 & G3 & \\
\midrule
Top-gated base & 55 & 8 & 14 & 15 & 18 & 12 & 18 & 19 & 6 & 18 & 19 & 18 & 0 \\
Hole and architecture probes & 19 & 0 & 12 & 4 & 3 & 13 & 6 & 0 & 0 & 12 & 4 & 3 & 15 \\
Gate/contact variants & 18 & 0 & 2 & 9 & 7 & 9 & 6 & 3 & 0 & 18 & 0 & 0 & 0 \\
Repeated devices and circuits & 18 & 0 & 0 & 7 & 11 & 0 & 6 & 9 & 3 & 2 & 11 & 5 & 0 \\
Three material roles & 12 & 0 & 6 & 5 & 1 & 0 & 0 & 7 & 5 & 3 & 7 & 2 & 0 \\
Scene/constraint variants & 6 & 0 & 5 & 1 & 0 & 1 & 2 & 3 & 0 & 0 & 5 & 1 & 0 \\
\midrule
Overall & 128 & 8 & 39 & 41 & 40 & 35 & 38 & 41 & 14 & 53 & 46 & 29 & 15 \\
\bottomrule
\end{tabular}
\endgroup

\end{table}

\section{Experimental setup}
\label{sec:protocol}
\paragraph{Workflows, scaffolds, and comparisons.}
We denote each condition as workflow/scaffold (e.g., ReAct-3/C0). In the base
scaffold (C0), each flake is represented by ordered contour vertices plus
harness-derived hints: a suggested gate-crossing point and source/drain line
segments. The agent constructs the gate and contact geometry from these
coordinates under the layout rules. C1 adds a reference point and direction
along each flake's medial axis (``spine''), and C2 adds text guidance to place two
contact rectangles on opposite sides of the crossing and extend the
gate beyond both flake boundaries. A deterministic interpreter applies the
planner's structured output. Plan-and-Execute (PaE) constructs the layout once,
without task-specific verifier feedback. ReAct-3 plans all layers jointly with
up to three planner rounds, including revision guided by verifier messages that
report expected and found transistor counts. ReAct-4, Reflexion-4, and ExpeL-4
construct the layout layer by layer with a code critic, a language-model fixer,
and layer locking (Appendix~\ref{app:protocol}). The workflow names refer to
our V1-derived variants inspired by ReAct, Reflexion, and ExpeL
\citep{yao2023react,shinn2023reflexion,zhao2024expel}. We evaluate six model
configurations (GPT5.6-Luna, GLM-5, Kimi K2.5, Qwen3-235B, Qwen3-32B, and
Ministral-3-14B) under ReAct-3/C0 on the same 128 tasks, with all five attempts
per task recorded. A second comparison evaluates GPT5.6-Luna across five
C0 workflows and ReAct-3 with scaffolds C1 and C2, also with five attempts per
task. The two comparisons share the ReAct-3/C0 baseline and contain
7,680 attempts. We chose ReAct-3/C0 on the 128 tasks as the primary
comparison after the workflow runs were complete, because it had complete,
comparable records for all six configurations. An earlier plan used ReAct-4/C0
on a 74-task set (Appendix~\ref{app:selection}). GPT5.6-Luna, GLM-5, and Kimi
K2.5 request high reasoning. The Bedrock Mantle releases of Qwen3-235B and
Ministral-3-14B are non-thinking or instruct models, and Qwen3-32B runs with
reasoning disabled (Appendix~\ref{app:protocol}). The panel therefore compares
model-and-serving configurations. All seven GPT5.6-Luna configurations use the
same API route, high reasoning, and a 16,000-token output cap. Historical GLM-5 workflow runs and
incomplete model campaigns appear in Appendix~\ref{app:historical}, and
Appendix~\ref{app:protocol} gives each configuration's full prompts and serving route.

\paragraph{Metrics and expert review.}
We report pass@1, pass@5, and all-five as defined in Equation~\ref{eq:passk}.
Marginal 95\% intervals use 10,000 task-bootstrap resamples, keeping all five
attempts of each resampled task. Within-GPT5.6-Luna pass@1 contrasts resample
104 clusters of tasks that share a specification apart from geometry and report simultaneous intervals over the six contrasts, and
Appendix~\ref{app:statistics} gives the resampling details. A senior
macro-device researcher with cleanroom fabrication experience in 2D-material
devices reviewed verifier-passing layouts against their task requirements on six
criteria: (1) the layout follows the requested device specification, (2) the
required FETs form on the intended flakes or regions, (3) each gate crosses the
intended channel at a physically meaningful position, (4) the contacts give
working source and drain access, (5) circuit tasks have the requested device
connections and circuit structure, and (6) the structure as a whole is suitable
for experimental use. The reviewer accepted a layout only when it had the
intended device structure and was judged experimentally usable under the
intended device interpretation. The audit covers GPT5.6-Luna, GLM-5, and Kimi
K2.5 under C0, the three ReAct-3/C0 configurations with the highest task
coverage, and GPT5.6-Luna under C1 and C2. For every task with any
verifier-passing attempt, we drew one of its passing layouts at random (436
task--configuration pairs in total), giving each covered task equal weight. Pass@1
and pass@5 keep their automated verifier definitions.

\section{Results}
\label{sec:results}
\subsection{Coverage and observed consistency}

\begin{table}[t]
\caption{Overall success and pass@1 by task property for six ReAct-3/C0
configurations on 128 tasks, with five attempts per task. Lower-block rates use
the $5n$ attempts of each subset of $n$ tasks, and subsets from different rows
overlap. 95\% intervals and the full breakdown are in
Appendix~\ref{app:demand}.}
\label{tab:performance}\centering\footnotesize
\begingroup\setlength{\tabcolsep}{3.5pt}\setlength{\abovetopsep}{4pt}
\begin{tabular}{@{}lrrrrrr@{}}
\toprule
 & GPT5.6-Luna & GLM-5 & Kimi K2.5 & Qwen3-235B & Qwen3-32B & Ministral-3-14B \\
\midrule
\multicolumn{7}{@{}l}{\emph{Overall (128 tasks, five attempts each)}} \\
pass@1 (\%) & \textbf{62.3} & 28.8 & 23.0 & 3.8 & 1.3 & 0.8 \\
pass@5 (\%) & \textbf{80.5} & 49.2 & 42.2 & 9.4 & 3.1 & 2.3 \\
All-five (\%) & \textbf{43.8} & 9.4 & 4.7 & 0.8 & 0.8 & 0.0 \\
\addlinespace[3pt]
\multicolumn{7}{@{}l}{\emph{Pass@1 (\%) by task property ($n$ tasks)}} \\
L1 (8) & \textbf{100.0} & 80.0 & 65.0 & 7.5 & 0.0 & 2.5 \\
L2 (39) & \textbf{91.8} & 55.4 & 44.6 & 10.3 & 4.1 & 2.1 \\
L3 (41) & \textbf{62.0} & 20.5 & 15.1 & 0.5 & 0.0 & 0.0 \\
L4 (40) & \textbf{26.5} & 1.0 & 1.5 & 0.0 & 0.0 & 0.0 \\
Without holes (113) & \textbf{59.8} & 26.2 & 20.5 & 3.0 & 1.2 & 0.9 \\
With holes (15) & \textbf{81.3} & 48.0 & 41.3 & 9.3 & 1.3 & 0.0 \\
No circuit (102) & \textbf{68.4} & 34.3 & 27.8 & 4.5 & 1.6 & 1.0 \\
Circuit (26) & \textbf{38.5} & 6.9 & 3.8 & 0.8 & 0.0 & 0.0 \\
\bottomrule
\end{tabular}
\endgroup

\end{table}

\begin{figure}[t]
\centering\includegraphics[width=.9\textwidth,trim=0 10pt 0 1pt,clip]{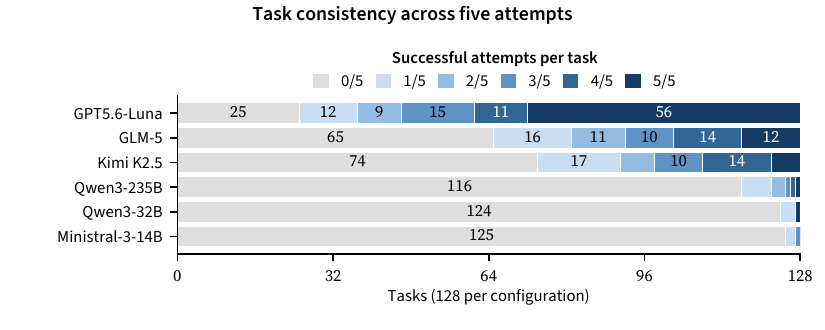}
\caption{Task-level consistency (ReAct-3/C0). Each bar splits the 128 tasks by
successes in five attempts. Pass@5 is the share outside 0/5, all-five is the
5/5 share, and pass@1 is the mean of $c_t/5$.}
\label{fig:distribution}
\end{figure}

Table~\ref{tab:performance} reports the six ReAct-3/C0 configurations, and
Appendix~\ref{app:demand} gives counts and intervals. GPT5.6-Luna succeeds on 399
of 640 attempts (62.3\% pass@1). It solves 103 of 128 tasks at least once
(80.5\% pass@5) and 56 in all five attempts (43.8\%). GLM-5 succeeds on 184 of
640 attempts (28.8\% pass@1, 49.2\% pass@5) and Kimi K2.5 on 147 (23.0\% and
42.2\%). Qwen3-235B, Qwen3-32B, and Ministral-3-14B succeed on fewer than 4\%
of their attempts and solve 12, 4, and 3 tasks at least once. Figure~\ref{fig:distribution} shows how
coverage and consistency differ. GPT5.6-Luna solves 56 tasks in all five
attempts and 47 in one to four attempts, and 25 tasks have no success. GLM-5
covers 63 tasks, 12 of them in all five attempts, and Kimi K2.5 covers 54, six
of them in all five. Together the six configurations solve 105 of 128 tasks.
Within this panel, GPT5.6-Luna uniquely solves 40 tasks and GLM-5 uniquely
solves two, and every task solved by the other four configurations is also
solved by GPT5.6-Luna. Across all 12 evaluated configurations, 117 tasks are
solved at least once.

\subsection{Performance by task property}
\begin{figure}[t]
\centering\includegraphics[width=.9\textwidth,trim=0 7pt 0 3pt,clip]{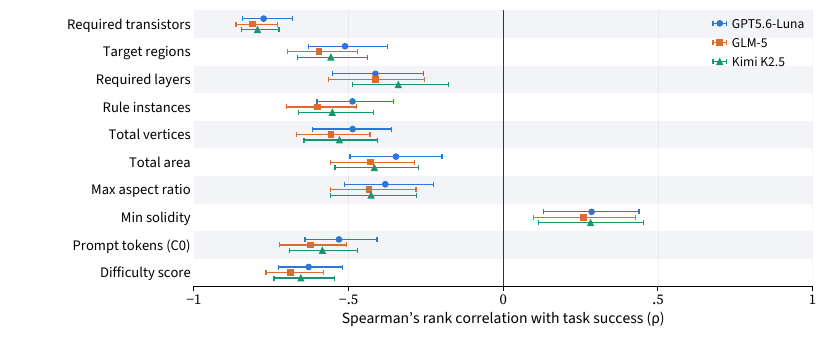}
\caption{Task-property associations with success under ReAct-3/C0. Points show
Spearman's rank correlation ($\rho$) between each task property and the number
of successful attempts out of five for GPT5.6-Luna, GLM-5, and Kimi K2.5. Bars
show 95\% bootstrap intervals over 128 tasks (127 for required transistors).
Negative values mean higher property values coincide with lower success.}
\label{fig:property-success}
\end{figure}
\begin{figure}[htb]
\centering\includegraphics[width=.78\textwidth]{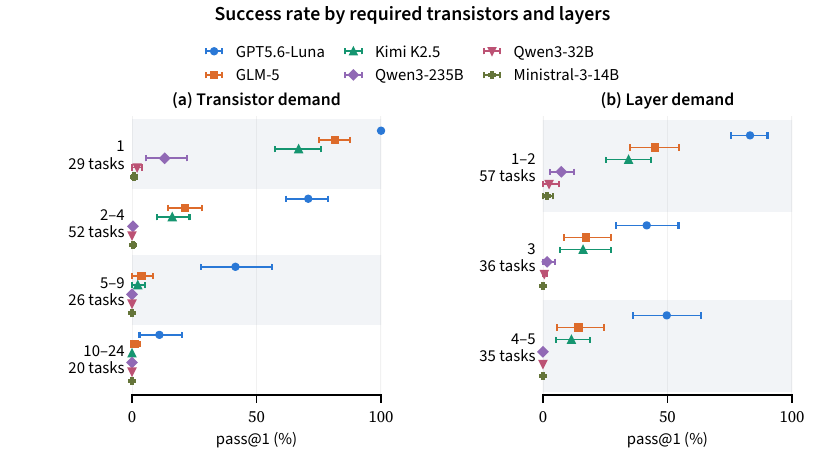}
\caption{Pass@1 by (a) required transistors per task and (b) editable design
layers for the six ReAct-3/C0 configurations, with 95\% task-bootstrap
intervals. Each bin label gives its task count.}
\label{fig:demand}
\end{figure}

The three configurations with pass@1 above 4\% perform best on L1. GPT5.6-Luna
reaches 100.0\% pass@1 on L1 and 26.5\% on L4 (Table~\ref{tab:performance}). The
corresponding rates are 80.0\% and 1.0\% for GLM-5 and 65.0\% and 1.5\% for
Kimi K2.5. GPT5.6-Luna reaches 38.5\% on circuit tasks and 68.4\% on other
tasks, and tasks with holed flakes have higher pass@1 than tasks without
(81.3\% versus 59.8\%). In Figure~\ref{fig:property-success}, required
transistor count has the strongest negative association with success ($\rho$
from $-0.77$ to $-0.81$). Every other property except minimum solidity is also
negatively associated with success, and minimum solidity has a weak positive
one ($\rho\approx0.26$--$0.29$). The figure omits Qwen3-235B,
Qwen3-32B, and Ministral-3-14B, which succeed on fewer than 4\% of attempts, and
Appendix~\ref{app:demand} gives all task strata with their 95\% intervals and
task counts. By transistor count, GPT5.6-Luna succeeds on 145 of 145 attempts on
one-transistor tasks, 184 of 260 on tasks needing 2--4, 54 of 130 on 5--9, and
11 of 100 on ten or more (Figure~\ref{fig:demand}). The one task with no
specified count succeeds 5/5 and is left out of these bins. The C0 hint
suggests one crossing per flake, and in 59 tasks at least one flake needs more
than one transistor. Among tasks needing 2--4 transistors, GPT5.6-Luna succeeds
on 87.7\% of attempts when each flake needs one transistor (26 tasks) and on
53.8\% when at least one flake needs more (26 tasks). For 5--9 transistors, the
rates are 66.0\% (10 tasks) and 26.2\% (16 tasks). In both demand groups,
success is higher when each flake needs one transistor. Success also falls as
the number of flakes grows. GPT5.6-Luna reaches 83.7\% pass@1 on the 49
single-flake tasks, 62.6\% on tasks with two or three flakes, 38.3\% with four
or five, and 32.9\% with six or more. GLM-5 and Kimi K2.5
reach 54.3\% and 43.7\% on single-flake tasks.

\subsection{GPT5.6-Luna workflow and scaffold comparisons}
Within GPT5.6-Luna,
ReAct-3/C0 succeeds more often than the four other C0 workflows. It exceeds PaE
by 27.0 pass@1 points and uses 2.46 times as many tokens. The simultaneous
intervals support the ReAct-3/C0 advantage over all four alternatives (Appendix
Figure~\ref{fig:intervention}). PaE reaches 35.3\% pass@1 and 50.8\% pass@5.
ExpeL-4 reaches 29.4\% and 47.7\%, Reflexion-4 25.5\% and 38.3\%, and ReAct-4
23.6\% and 38.3\%. ReAct-3/C0 exceeds these three layer-by-layer workflows by
33.0--38.8 points, and they use 4.92--5.54 times its tokens. C1 reaches 67.0\% pass@1, 85.9\% pass@5, and 46.1\%
all-five, the highest observed values among the seven configurations. C1 differs from C0 by $+4.7$
pass@1 points [$-0.8$, 10.1] with 0.99 times the tokens, and C2 differs by
$+0.3$ points [$-5.0$, 5.6] with 1.19 times the tokens. Both contrast intervals include
zero. Token counts include failed attempts and are compared only within
GPT5.6-Luna because tokenizers differ. Appendix Table~\ref{tab:luna} reports
success metrics, token totals, and completed agent calls for all seven
configurations.

\subsection{Observed failure symptoms}
\label{sec:failures}
Transistor-count failures are the most common failure type for all six
configurations. They appear in 191 of GPT5.6-Luna's 241 failed attempts
(Appendix Figure~\ref{fig:failure-primary}a). A failed layout often fails
several checks at once, and a layout with an empty required layer may
also fail crossings, overlaps, and transistor counts. An agent can also
declare a task \texttt{INFEASIBLE}, although every task has a passing reference
layout. The declaration ends the attempt and leaves required layers empty. It
accounts for 63 of GPT5.6-Luna's 241 failed attempts and 573 of
Ministral-3-14B's 635. For GPT5.6-Luna, 63 of its 64 empty-layer failures and all 63
crossing failures are in attempts with a declaration (Appendix
Table~\ref{tab:systematic-trajectories}). Qwen3-235B and Qwen3-32B fail the
saved-layout input-preservation check in 32 and 6 of their 640 attempts, and the
other four configurations pass it in every attempt. GPT5.6-Luna has 223
successes at the first verifier check and 176 after revision. GLM-5 has 73 and
111, and Kimi K2.5 has 48 and 99, so revision rounds account for 44\%, 60\%,
and 67\% of their successes. Most of their failures end at the round cap: 178
for GPT5.6-Luna, 421 for GLM-5, and 460 for Kimi K2.5. At the first, second, and
third checks, 223 of 578, 126 of 354, and 50 of 228 checked GPT5.6-Luna layouts
pass (Appendix Table~\ref{tab:systematic-rounds}). Later checks include only
attempts that are still unresolved, and the pass rate among checked layouts
falls from 38.6\% at the first check to 35.6\% and 21.9\%. Revision logs also
track changes to individual checks. Across 4,677 consecutive pairs of
verifier checks, 1,390 transitions resolve at least one check type while introducing
another (Appendix~\ref{app:systematic-errors}).

\subsection{Expert review of verifier-passing layouts}
\label{sec:expert-review}
\begin{figure}[htb]
\centering\includegraphics[width=\textwidth]{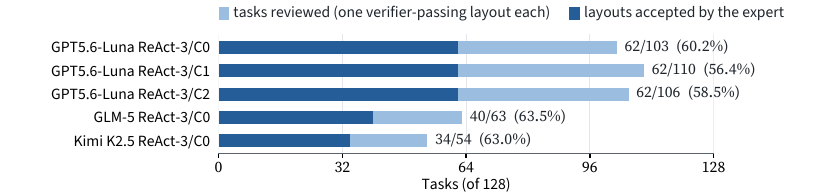}
\caption{Expert review of one randomly sampled verifier-passing layout per
covered task. Light bars show reviewed task--configuration pairs, and dark bars
show layouts the expert accepted.}
\label{fig:expert-review}
\end{figure}
In Figure~\ref{fig:expert-review}, the expert accepted
62 of 103 layouts for GPT5.6-Luna ReAct-3/C0 (60.2\%), 62 of 110 for C1 (56.4\%),
and 62 of 106 for C2 (58.5\%). For GLM-5 and Kimi K2.5 under ReAct-3/C0, the
counts are 40 of 63 (63.5\%) and 34 of 54 (63.0\%). In total, the expert accepted 260 of the 436
reviewed layouts (59.6\%), and acceptance ranges from 56.4\% to 63.5\% across
the five configurations. These rates describe expert judgments of one sampled
passing layout per covered task, and all other attempts keep their automated
scores. Each rate applies to its own configuration's covered task set, so
comparing configurations would need a common set of reviewed tasks.

\section{Discussion and conclusion}
\benchmark{} V2 provides 128 executable layout tasks built on microscopy-derived
flake contours. It lets researchers test whether language-model agents can build
flake-specific device layouts from task text and contour coordinates, and how
reliably they do so across five attempts on the same fixed input. Because each
task gives only text and coordinates, success requires spatial reasoning from a
textual device specification to flake-specific layout geometry. Across the 12
evaluated configurations, agents solve 117 of the 128 tasks at least once.
GPT5.6-Luna solves 103
tasks at least once and 56 in all five attempts, and the six-model panel covers
105 tasks. Within GPT5.6-Luna, ReAct-3/C0 exceeds PaE by 27.0 pass@1 points at
2.46 times the recorded tokens. The expert audit adds a device-level check: across the five audited
configurations, the expert accepts 56.4\%--63.5\% of sampled verifier-passing
layouts, or 260 of 436 in total. The results suggest two ways to improve layout
agents: match gate and contact construction to the transistor demand on each
flake, and revise layouts without breaking constraints that already hold. The agentic workflow also shows potential for research acceleration: with eight tasks
running in parallel, GPT5.6-Luna with ReAct-3/C0 completes one attempt on all 128 tasks in
about 1.6 hours, the longest of the eight parallel first-attempt runs in our
records. Based on our prior layout-design experience, we estimate that manually
constructing layouts for all 128 tasks would take about 20 hours. This suggests that
agents can build device layouts about ten times faster, if their layouts are as
valid as manual ones at the device level. Such a speedup would shorten device design
cycles and speed up 2D-material device research.

\label{main:text:end}\label{main:end}

\clearpage
\subsection*{AI use statement}
In this work, we used generative AI tools to develop code for task
construction, verification, and analysis, to support qualitative analysis of
agent traces and task specifications, to give feedback on the methodology, and
to interpret results. We have not used generative AI tools to generate the
benchmark source data or the expert-review labels. Flake contours derive from
released microscopy masks, multi-flake scenes come from our sampling code, the
evaluated layouts are outputs of the models under study, and a human expert
performed the expert review. Mathematical proofs and translation are not
applicable to this work. Additionally, we used generative AI tools to edit
figures and text, draft parts of the paper, and search and summarize the
literature. We have reviewed all AI-assisted work, and we take responsibility
for the final content of this work, including text, claims, and artifacts
produced with the aid of generative AI.

\subsection*{Ethics statement}
Task contours derive from segmentation masks in the MaskTerial
datasets, released under the MIT license
\citep{maskterial_dataset,uslu2025maskterial}. One of these datasets reuses the
images of the 2DMatGMM datasets, released under CC BY 4.0 \citep{uslu2023twodmatgmm}.
Appendix~\ref{app:datacard} records provenance and license information. The
expert review judged generated layouts and did not involve human-subject data.
The benchmark evaluates layout geometry and structure, and electrical operation
and fabrication readiness need separate validation.

\subsection*{Reproducibility statement}
The task construction, scoring rules, analysis inputs, source mappings,
uncertainty methods, and evaluated prompt configuration are documented in the
appendices, with the prompt configuration in Appendix~\ref{app:protocol}. An
evaluator-only package preserves the active task bytes and scoring code, adds
fixed-input checking, and has passed a clean-install test and selected fixtures
(Appendix~\ref{app:release}). The data card is in Appendix~\ref{app:datacard}.
The code and the dataset, including tasks, verifier, and reference layouts,
will be released publicly. These materials support rescoring layouts and reproducing analyses
from recorded outcomes. Exact replay of provider-side runs depends on access to
the historical hosted model revisions.
\clearpage
\bibliography{refs}
\bibliographystyle{plainnat}
\clearpage
\appendix
\FloatBarrier
\section{Statistical methods and multiplicity sensitivity}
\label{app:statistics}
The contrasts use ReAct-3 at the base scaffold (C0) as their reference.
The marginal endpoint intervals use the completed 10,000-draw task bootstrap
(seed 20260920). A draw selects 128 task indices with replacement and retains
the five outcomes within each task. Every arm receives the same indices;
paired differences are calculated within each draw. Pass@5 uses the observed
five-attempt outcomes. Each trajectory index denotes a fresh run; matching
indices across arms do not indicate shared random seeds. Subgroup intervals use 4,000 within-group task-bootstrap draws
(seed 20260922 in the original calculation), with common draws across models. Task and cluster resampling both vary the observed composition.

For the cluster-resampling analysis used to construct simultaneous intervals,
let a group $g$ contain $n_g$ tasks, with summed
endpoint contributions $U_{ag}$ for arm $a$. Draw $G$ groups with replacement
and estimate an arm's endpoint by
\begin{equation}
 \widehat\theta_a^*=\frac{\sum_{b=1}^{G}U_{a,g_b}}{\sum_{b=1}^{G}n_{g_b}}.
\end{equation}
This estimates the task-weighted endpoint even when cluster sizes differ. All arms and
endpoints share each of the 50,000 draws (seed 20260921). Groupings are 128
individual tasks, 104 specification groups, 82 source-image groups, or 61
connected image-or-specification groups (largest component: 31 tasks).
Specification groups share a signature of architecture, layer table, design
order, DRC rules, hidden structural criteria, and per-region material, type,
and hole status. Source-image groups are connected components of the task
graph joined by shared source-image keys; the combined grouping also joins
identical specifications. We define these groups from stored task metadata.
The 82 and 61 image-based clusters use dataset-prefixed source keys.
Physical-specimen independence would require source-image deduplication;
the primary sensitivity analysis uses specification clusters.

For contrast $j$, $s_j$ is the standard deviation of its bootstrap estimates.
For a declared family $\mathcal J$, the critical value is
\begin{equation}
 c_{.95}=Q_{.95}\!\left(\max_{j\in\mathcal J}
 \left|\frac{\widehat\delta_j^*-\widehat\delta_j}{s_j}\right|\right),
 \qquad I_j=\widehat\delta_j\pm c_{.95}s_j.
\end{equation}
These are simultaneous intervals from \emph{fixed-SE standardized bootstrap
deviations}, using one standard-error estimate per contrast across resamples.
Their coverage is a bootstrap approximation. The choice to emphasize
pass@1 was made retrospectively. Six contrasts within each endpoint
form three separate adjustment families. Joint coverage of all 18 findings
uses the additional all-18 adjustment.
Table~\ref{tab:stat-pass1} distinguishes unadjusted intervals, six-contrast
endpoint-specific adjustment, and joint all-18 adjustment.

\begin{table}[h]\centering\small
\caption{Pass@1 differences versus GPT5.6-Luna ReAct-3/C0, in percentage points.
C1 and C2 are ReAct-3 scaffold variants; workflow alternatives use C0.
Specification-cluster intervals use the completed joint analysis.}
\label{tab:stat-pass1}\begin{tabular}{lrrrr}
\toprule
Contrast & $\Delta$ (pp) & Unadjusted & Six pass@1 & All 18 \\
\midrule
C1: + position hints & 4.69 & [0.56, 9.00] & [-0.75, 10.13] & [-1.50, 10.87] \\
C2: + structural guidance & 0.31 & [-3.81, 4.39] & [-4.97, 5.59] & [-5.69, 6.32] \\
Plan-and-Execute & -27.03 & [-33.69, -21.17] & [-35.15, -18.92] & [-36.26, -17.80] \\
ExpeL-4 & -32.97 & [-39.72, -26.31] & [-41.63, -24.31] & [-42.82, -23.11] \\
Reflexion-4 & -36.88 & [-43.53, -30.57] & [-45.24, -28.51] & [-46.39, -27.36] \\
ReAct-4 & -38.75 & [-45.41, -31.58] & [-47.66, -29.84] & [-48.89, -28.61] \\
\bottomrule
\end{tabular}
\end{table}

The exploratory pass@5 and all-five contrasts, each adjusted within its own
six-contrast family, follow the same pattern. All four alternative workflows
are lower than ReAct-3/C0 (pass@5: $-29.7$ to $-42.2$ points; all-five:
$-28.1$ to $-35.2$), with every six-contrast interval excluding zero. The
scaffold contrasts stay near zero, and every C1 and C2 interval spans zero
(pass@5 C1 $+5.5$, $[-2.2, 13.2]$; all-five C1 $+2.3$, $[-4.2, 8.9]$).

C1's unadjusted specification-cluster pass@1 interval is $[0.56,9.00]$ points;
the corresponding percentile Bonferroni interval is $[-0.79,10.60]$.
C1's pass@1 intervals, adjusted over six contrasts, span zero under task,
specification, image, and connected groupings (Table~\ref{tab:cluster}). These intervals retain uncertainty about the direction and size of the C1 difference.
The complete numerical export also retains Bonferroni bounds and secondary
endpoints for each grouping.
\begin{table}[h]\centering\small
\caption{C1 pass@1 simultaneous interval under alternative groupings;
adjustment is over six pass@1 contrasts in every row.}
\label{tab:cluster}\begin{tabular}{lrrr}
\toprule
Grouping & Clusters & Largest & C1 $\Delta$ pass@1 interval \\
\midrule
Task & 128 & 1 & [-0.42, 9.80] \\
Specification & 104 & 9 & [-0.75, 10.13] \\
Source image & 82 & 17 & [-0.54, 9.92] \\
Image/spec connected & 61 & 31 & [-0.95, 10.33] \\
\bottomrule
\end{tabular}
\end{table}

\paragraph{Variation across repeated runs.}
Reweighting designed tasks asks how summaries respond to task composition.
On the fixed task set a plug-in repeated-run standard error is
$\sqrt{\sum_t \widehat p_t(1-\widehat p_t)/4}/128$, with
$\widehat p_t=c_t/5$, under independent stable task-level attempts. With five trials, tasks that always pass or always fail contribute zero
estimated variance, so future-run uncertainty can be understated. All-five
records consistency in those five attempts. The composition intervals describe
the observed tasks, including their reused sources, selected templates, and
feasibility screening. They condition on each configuration's recorded source
revision and serving settings. ExpeL also depends on task sequence and shard
assignment; estimating its variation across new sequences requires repeated
sequence-level runs.

\FloatBarrier
\section{Configuration and resource details}
\label{app:protocol}
The base scaffold (C0) suppresses the dedicated spine section, but derived crossing points and
source/drain zones still appear in active constraints. C1 adds a spatial
reference point and spine direction for each region, with contacts aligned
along the spine and a perpendicular gate. C2 adds structural guidance: two
separate contact rectangles on opposite sides of the crossing, a gap computed
from gate width plus twice the minimum separation, and a gate extending beyond
both flake boundaries. The scaffolds supply different guidance. Reference constructions remain with
the evaluator in all three conditions.

\paragraph{Implemented workflows and scaffolds.}
ReAct-4, Reflexion-4, and ExpeL-4
use layer-wise planning, a code Critic, an LLM Fixer, locking, and deferred
checks. All three invoke a Reflector, with additional successful-layer
reflection in Reflexion-4 and ExpeL-4. ExpeL appends
experience records after tasks and exposes the three most recent records to
subsequent planning within each shard/attempt sequence. Its outcomes therefore
depend on task order and shard assignment.

Every first planner prompt includes the objective, rounded contour vertices,
layer table, design order, layout rules, and derived active constraints,
which give one suggested crossing per flake. The fixed system prompt describes
a two-contact transistor and asks the planner to follow a prescriptive layout
plan; at C0 that plan section is empty. Layer-wise planners
also receive a one-transistor contact exemplar; whole-layout workflows do not.
Thus prompt content, verification timing, and construction opportunities differ
between workflow packages. The guarded drawing tools can repair
self-intersections by convex hull, and the per-layer harness implements
conditional contact completion. The six-model ReAct-3 panel uses direct interpretation of planner operations;
its C0 prompts supply spine-derived geometry hints.

\begin{table}[h]\centering\small
\caption{Complete GPT5.6-Luna arms: 640 attempts each. C1 and C2 are ReAct-3
scaffold variants; the other rows use C0. Endpoints are percentages;
M tokens denotes millions of recorded input plus output tokens, including
failures. Agent runs are completed \texttt{agent.run()} calls.}
\label{tab:luna}\begin{tabular}{lrrrrrr}
\toprule
Arm & Successes & pass@1 & pass@5 & All-five & M tokens & Agent runs \\
\midrule
ReAct-3/C0 & 399 & 62.3 & 80.5 & 43.8 & 13.56 & 1223 \\
C1: + position hints & 429 & 67.0 & 85.9 & 46.1 & 13.46 & 1200 \\
C2: + structural guidance & 401 & 62.7 & 82.8 & 43.0 & 16.17 & 1357 \\
Plan-and-Execute & 226 & 35.3 & 50.8 & 15.6 & 5.50 & 640 \\
ExpeL-4 & 188 & 29.4 & 47.7 & 13.3 & 75.15 & 8591 \\
Reflexion-4 & 163 & 25.5 & 38.3 & 9.4 & 67.40 & 8409 \\
ReAct-4 & 151 & 23.6 & 38.3 & 8.6 & 66.72 & 7502 \\
\bottomrule
\end{tabular}
\end{table}

The C0 baseline records 13,558,139 tokens and 1,223 agent runs; PaE records
5,503,519 tokens and 640 runs. Token ratios divide arm totals by baseline totals,
not by successes. Cross-model ratios and dollar estimates are not reported.
Provider-internal HTTP requests, retry lineage, stable hosted revisions, and
billed costs cannot be reconstructed from these counters. Matching current
prompt modules and task hashes does not establish identical generation-time
harness revisions for older campaigns. In particular, the five Bedrock Mantle runs in the six-model panel lack
a complete generation-time harness hash record. In the six-model panel,
GPT5.6-Luna uses the OpenAI Responses API and the other five use Bedrock Mantle
chat completions; GPT5.6-Luna, GLM-5, and Kimi K2.5 request high reasoning,
Qwen3-235B and Ministral-3-14B use non-thinking/instruct releases, and
Qwen3-32B has reasoning disabled. Output-validation settings also differ.

\paragraph{Within-GPT5.6-Luna execution differences.}
All seven configurations request GPT5.6-Luna via the Responses API, high
reasoning, a 16,000-token output cap, 12 output-validation retries, and a
900-second timeout. The ReAct-3/C0 baseline records 60 requests per minute and
eight concurrent workers; the six later arms record 90 and twelve. Recorded
orchestrator and agent source hashes differ, while prompt-module and verifier
hashes agree. Archived hashes document these differences; a complete execution comparison
would require the remaining source versions. The results compare implemented
configurations on matched tasks, with each configuration's recorded resource use.

\paragraph{Reflection and accumulated experience.}
ReAct-4 invokes a post-layer reflector; Reflexion-4 and ExpeL-4 additionally
have a successful-Critic-to-Reflector graph edge and can reflect twice after
a successful layer. The configurations therefore differ in both reflection timing and frequency. ExpeL writes a JSON experience record after each task and supplies
the last three records through its planning-context builder. The completed GPT5.6-Luna
run has 40 shard/attempt databases with 640 entries, and logs confirm writes.
These records verify accumulation and the mechanism that supplies experience
to planning. Retrieval uses recency. Verifying the exact text consumed in each
historical prompt would require a complete prompt archive. The database is scoped to each
shard/attempt sequence; task order and shard assignment affect subsequent inputs.

\paragraph{Choice of the primary panel.}
\label{app:selection}
An earlier internal protocol document, dated August 6 and marked frozen,
designated ReAct-4/C0 on the 74-task residual set, scored by pass@5, as the
main benchmark condition. It served as an internal experiment plan. ReAct-4/C0 runs on these tasks were carried out for
several models but are not the primary comparison reported here. A September 5
fix enforced ReAct-3's three-round bound and restored the initial scene between
plans; older runs could exceed that bound. A follow-up planning note of
September 6 kept ReAct-4/C0 as the main condition and treated ReAct-3 on the 128
tasks as exploratory, as did status notes through September 15. By September
18, the complete ReAct-3 panel became primary because its common-task records were comparable
across models, unlike the wider matrix, whose serving and harness conditions
differed. The change followed the availability of comparative outcomes, so the
selection was retrospective. The records establish this sequence of events
but leave the exact decision date and the role of performance unspecified.
The later GPT5.6-Luna comparison evaluates the tested configurations on the
128-task set.

\FloatBarrier
\section{GPT5.6-Luna workflow and scaffold comparisons}
\label{app:workflows}
\begin{figure}[t]
\centering\includegraphics[width=\textwidth]{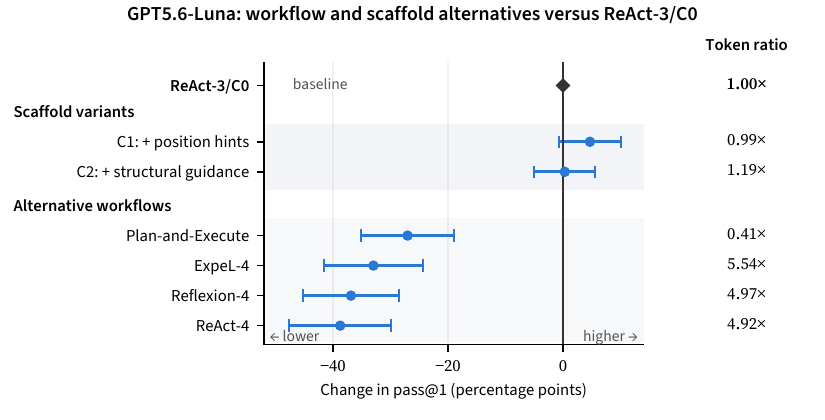}
\caption{GPT5.6-Luna pass@1 differences and token ratios for each alternative
versus ReAct-3/C0 (640 attempts per arm); positive differences favor the
alternative. Intervals are retrospective 95\% simultaneous intervals over the
six pass@1 contrasts, from paired specification-cluster resampling with
fixed-SE standardized deviations. Contrasts compare task-matched
configurations, not isolated mechanisms, and retain source-revision,
concurrency, and ExpeL-sequence differences.}
\label{fig:intervention}
\end{figure}

C1 changes pass@1 by $+4.69$ points, with simultaneous interval
$[-0.75,10.13]$, at $0.99\times$ C0 tokens. The simultaneous interval spans zero. C2 changes pass@1 by
$+0.31$ points, $[-4.97,5.59]$, at $1.19\times$ tokens. Both intervals span zero; the analysis estimates
differences in success rates rather than testing equivalence. Adjustment across all 18 combinations
widens the C1 interval to $[-1.50,10.87]$ points.

Tables~\ref{tab:stat-pass1} and~\ref{tab:cluster} separate grouping from
adjustment: C1's unadjusted specification-cluster interval is positive, but its
six-contrast intervals span zero under every grouping, as do all C2 intervals. Image-based clusters use provenance keys; source overlap is documented in
Appendix~\ref{app:construction}.

\paragraph{Observed infeasibility declarations.}
GPT5.6-Luna ReAct-4, Reflexion-4, and ExpeL-4 declare \texttt{INFEASIBLE} in
405/640, 388/640, and 350/640 attempts, respectively (63.3/60.6/54.7\%).
First declarations occur after the first designed layer in 397/405, 376/388,
and 345/350 declaring attempts; contacts are not always first. Locking, deferred
checks, and spine constraints are code-supported observational explanations,
not isolated mechanisms or adjudicated causes. Appendix~\ref{app:infeasible}
gives stages, repeated declarations, and selected traces.

On the 69 via tasks, ReAct-4, Reflexion-4, and
ExpeL-4 pass 4, 4, and 14 of 345 attempts, respectively; ReAct-3 passes 152. In
296/333 ReAct-4 via-task attempts with contacts, the saved contacts cannot
enclose a via at the visible $0.5\,\mu$m margin (Reflexion-4: 300/329;
ExpeL-4: 294/339). The per-layer Critic defers cross-layer checks until the
via is designed, after contacts are locked. This pattern is consistent with an effect of the contact exemplar or the
verification schedule. The data do not separate the two
(Appendix~\ref{app:infeasible}).

\FloatBarrier
\section{Infeasibility census and selected observations}
\label{app:infeasible}
The declaration census covers 3,840 logs from the six configurations in
Table~\ref{tab:census}, a subset of the 7,680 current-panel attempts. An
independent recheck confirmed all paths and declaration counts. A declaring attempt may contain multiple declarations, and an
\texttt{INFEASIBLE} message is not an independently validated infeasibility
proof. Workflow-level declarations are recorded in logs, separately from
saved-layout verdicts. Table~\ref{tab:census} separates the denominators; Table~\ref{tab:stages}
reports the named stage of the first declaration in the GPT5.6-Luna layer-wise arms.

\begin{table}[h]\centering\small
\caption{Infeasibility-declaration census. Counts of attempts and declarations are distinct.}
\label{tab:census}\begin{tabular}{lrrr}
\toprule
Arm & Attempts & Declaring attempts & Declarations \\
\midrule
Historical GLM ReAct-4 & 640 & 164 & 260 \\
GPT5.6-Luna ExpeL & 640 & 350 & 428 \\
GPT5.6-Luna PaE & 640 & 58 & 58 \\
GPT5.6-Luna ReAct-3 & 640 & 63 & 63 \\
GPT5.6-Luna ReAct-4 & 640 & 405 & 515 \\
GPT5.6-Luna Reflexion & 640 & 388 & 504 \\
\bottomrule
\end{tabular}
\end{table}
\begin{table}[h]\centering\small
\caption{First-declaration stages. Contacts account for 19/20/9 attempts; the final
column gives declarations on layers following contacts. Contacts are not always
the first designed layer; the following paragraph distinguishes the two temporal
definitions.}
\label{tab:stages}\begin{tabular}{lrrrrr}
\toprule
Arm & Contacts & Gate/top gate & Via & Pad cover & After contacts \\
\midrule
ReAct-4 & 19 & 169 & 217 & 0 & 386/405 \\
Reflexion-4 & 20 & 145 & 223 & 0 & 368/388 \\
ExpeL-4 & 9 & 120 & 220 & 1 & 341/350 \\
\bottomrule
\end{tabular}
\end{table}

Matching each first declaration to its task's design order gives 8/12/5 on
the first designed layer and 397/376/345 on later layers, respectively.
By contrast, 386/368/341 occur after contacts: in some architectures a
bottom/back gate precedes contacts. The workflow comparison uses the design-order
definition; neither denominator counts all repeated declarations.

\paragraph{Possible explanations supported by the implementation.}
The layer-wise Critic suppresses checks referencing layers not yet designed;
prior geometry can then be locked before later-layer conflicts are checked.
Prompts also prescribe spine-based constraints. A geometry screen examines saved layouts associated with
1,093 first gate/via declarations for possible geometric obstructions, and the detailed trace set contains
47 selected cases. Neither is a randomized intervention or an unbiased causal
sample. Bounding-box obstructions can support a particular enclosure failure;
a sufficiently large box or contact gap cannot establish complete geometric,
topological, and routing feasibility. Final saved layouts may also differ
from the state at declaration. We report no causal attribution percentage or
calibrated error margin from these screens.

\begin{table}[h]\centering\small
\caption{Selected trace evidence (case IDs from the 47-case audit).
Interpretations are deliberately limited to the observed condition.}
\label{tab:traces}
\begin{tabular}{p{.13\textwidth}p{.23\textwidth}p{.54\textwidth}}
\toprule
Case / arm & Task and attempt & Observation and interpretation \\
\midrule
C04 / R4 & L3/case12, run 4 & Via declaration after contacts were locked.
The recorded $0.5\times0.5\,\mu$m contacts cannot enclose a positive-area via
with $0.5\,\mu$m margins on both sides. This supports that local obstruction.\\
C01 / R4 & L1/case10, run 5 & Gate declaration with a prescribed crossing.
The saved-layout screen finds a $2.80\,\mu$m contact gap; this alone does not
prove a complete valid layout at the declaration state.\\
\bottomrule
\end{tabular}
\end{table}

\FloatBarrier
\section{Task demand and property dependence}
\label{app:demand}
\begin{table}[h]
\caption{Overall metrics with intervals and counts for the six ReAct-3/C0
configurations (128 tasks, five attempts each). pass@1 is the mean success
rate over 640 attempts; pass@5 the fraction of tasks solved at least once;
All-five the fraction solved in all five attempts. Values show percentage
[95\% task-bootstrap interval] with counts beneath. Intervals reflect
sensitivity to task composition; lower bounds of 0.0 reflect resampling
support, not an estimated zero chance of success. $^{\dagger}$No Ministral
task succeeds in all five attempts; the degenerate $[0.0,0.0]$ interval is
omitted.}
\label{tab:performance-full}\centering\small
\begin{tabular}{lrrr}
\toprule
Configuration & pass@1 & pass@5 & All-five \\
\midrule
\raisebox{3.5pt}{GPT5.6-Luna} & \shortstack[r]{\textbf{62.3\% [55.5, 69.2]}\\\scriptsize(399/640)} & \shortstack[r]{\textbf{80.5\% [73.4, 86.7]}\\\scriptsize(103/128)} & \shortstack[r]{\textbf{43.8\% [35.2, 52.3]}\\\scriptsize(56/128)} \\
\raisebox{3.5pt}{GLM-5} & \shortstack[r]{28.8\% [22.8, 35.0]\\\scriptsize(184/640)} & \shortstack[r]{49.2\% [40.6, 57.8]\\\scriptsize(63/128)} & \shortstack[r]{9.4\% [4.7, 14.8]\\\scriptsize(12/128)} \\
\raisebox{3.5pt}{Kimi K2.5} & \shortstack[r]{23.0\% [17.7, 28.8]\\\scriptsize(147/640)} & \shortstack[r]{42.2\% [33.6, 50.8]\\\scriptsize(54/128)} & \shortstack[r]{4.7\% [1.6, 8.6]\\\scriptsize(6/128)} \\
\raisebox{3.5pt}{Qwen3-235B} & \shortstack[r]{3.8\% [1.6, 6.4]\\\scriptsize(24/640)} & \shortstack[r]{9.4\% [4.7, 14.8]\\\scriptsize(12/128)} & \shortstack[r]{0.8\% [0.0, 2.3]\\\scriptsize(1/128)} \\
\raisebox{3.5pt}{Qwen3-32B} & \shortstack[r]{1.3\% [0.2, 3.1]\\\scriptsize(8/640)} & \shortstack[r]{3.1\% [0.8, 6.3]\\\scriptsize(4/128)} & \shortstack[r]{0.8\% [0.0, 2.3]\\\scriptsize(1/128)} \\
\raisebox{3.5pt}{Ministral-3-14B} & \shortstack[r]{0.8\% [0.0, 2.0]\\\scriptsize(5/640)} & \shortstack[r]{2.3\% [0.0, 5.5]\\\scriptsize(3/128)} & \shortstack[r]{0.0\%$^{\dagger}$\\\scriptsize(0/128)} \\
\bottomrule
\end{tabular}

\end{table}
\begin{table}[h]
\caption{Pass@1 (\%) [95\% task-bootstrap interval] by task stratum for the
six ReAct-3/C0 configurations; task counts in parentheses. L is the template
level, D the rule-based difficulty class, G the shape tier; holes,
circuit templates, and target-region counts are task properties. Strata
overlap in their covariates and are descriptive.}
\label{tab:stratified-full}\centering
\begingroup\setlength{\tabcolsep}{1.5pt}\scriptsize
\begin{tabular}{@{}lrrrrrr@{}}
\toprule
 & GPT5.6-Luna & GLM-5 & Kimi K2.5 & Qwen3-235B & Qwen3-32B & Ministral-3-14B \\
\midrule
\multicolumn{7}{@{}l}{\emph{Template level}} \\
L1 (8) & 100.0 [100.0, 100.0] & 80.0 [67.5, 92.5] & 65.0 [55.0, 72.5] & 7.5 [0.0, 17.5] & 0.0 [0.0, 0.0] & 2.5 [0.0, 7.5] \\
L2 (39) & 91.8 [86.7, 96.4] & 55.4 [43.6, 66.7] & 44.6 [32.8, 56.4] & 10.3 [3.6, 17.9] & 4.1 [0.5, 10.3] & 2.1 [0.0, 5.6] \\
L3 (41) & 62.0 [50.7, 73.2] & 20.5 [13.2, 28.3] & 15.1 [7.8, 22.9] & 0.5 [0.0, 1.5] & 0.0 [0.0, 0.0] & 0.0 [0.0, 0.0] \\
L4 (40) & 26.5 [16.5, 37.0] & 1.0 [0.0, 2.5] & 1.5 [0.0, 4.0] & 0.0 [0.0, 0.0] & 0.0 [0.0, 0.0] & 0.0 [0.0, 0.0] \\
\multicolumn{7}{@{}l}{\emph{Difficulty class}} \\
D1 (35) & 97.1 [94.3, 99.4] & 66.9 [56.6, 76.6] & 56.0 [45.7, 66.3] & 13.1 [5.7, 22.3] & 4.6 [0.6, 11.4] & 2.3 [0.0, 6.3] \\
D2 (38) & 68.4 [57.9, 78.9] & 25.3 [14.7, 36.8] & 18.9 [10.0, 28.9] & 0.5 [0.0, 1.6] & 0.0 [0.0, 0.0] & 0.5 [0.0, 1.6] \\
D3 (41) & 40.0 [28.3, 51.7] & 8.8 [4.4, 13.7] & 5.9 [1.5, 11.2] & 0.0 [0.0, 0.0] & 0.0 [0.0, 0.0] & 0.0 [0.0, 0.0] \\
D4 (14) & 24.3 [8.6, 42.9] & 1.4 [0.0, 4.3] & 1.4 [0.0, 4.3] & 0.0 [0.0, 0.0] & 0.0 [0.0, 0.0] & 0.0 [0.0, 0.0] \\
\multicolumn{7}{@{}l}{\emph{Shape tier}} \\
G1 (53) & 72.5 [62.3, 81.9] & 35.8 [26.0, 46.0] & 30.9 [21.9, 40.4] & 6.0 [1.5, 11.7] & 3.0 [0.4, 7.5] & 1.5 [0.0, 4.2] \\
G2 (46) & 56.5 [45.2, 67.8] & 25.2 [15.7, 35.2] & 15.2 [7.8, 23.5] & 2.2 [0.4, 4.3] & 0.0 [0.0, 0.0] & 0.0 [0.0, 0.0] \\
G3 (29) & 53.1 [37.9, 66.9] & 21.4 [11.7, 33.1] & 20.7 [9.0, 33.8] & 2.1 [0.0, 5.5] & 0.0 [0.0, 0.0] & 0.7 [0.0, 2.1] \\
\multicolumn{7}{@{}l}{\emph{Hole probes}} \\
Without holes (113) & 59.8 [52.0, 67.3] & 26.2 [20.0, 32.6] & 20.5 [15.0, 26.2] & 3.0 [0.7, 5.8] & 1.2 [0.0, 3.2] & 0.9 [0.0, 2.1] \\
With holes (15) & 81.3 [65.3, 94.7] & 48.0 [26.7, 68.0] & 41.3 [22.7, 61.3] & 9.3 [2.7, 17.3] & 1.3 [0.0, 4.0] & 0.0 [0.0, 0.0] \\
\multicolumn{7}{@{}l}{\emph{Circuit specification}} \\
No circuit (102) & 68.4 [61.0, 75.7] & 34.3 [27.1, 41.8] & 27.8 [21.4, 34.9] & 4.5 [1.8, 7.8] & 1.6 [0.2, 3.9] & 1.0 [0.0, 2.4] \\
Circuit (26) & 38.5 [24.6, 53.1] & 6.9 [1.5, 13.8] & 3.8 [0.8, 7.7] & 0.8 [0.0, 2.3] & 0.0 [0.0, 0.0] & 0.0 [0.0, 0.0] \\
\multicolumn{7}{@{}l}{\emph{Target regions}} \\
1 (49) & 83.7 [75.5, 91.0] & 54.3 [43.3, 64.5] & 43.7 [33.1, 54.3] & 9.4 [4.1, 15.9] & 3.3 [0.4, 7.8] & 1.6 [0.0, 4.5] \\
2--3 (39) & 62.6 [49.7, 74.9] & 21.5 [13.3, 30.3] & 19.0 [11.3, 27.7] & 0.5 [0.0, 1.5] & 0.0 [0.0, 0.0] & 0.5 [0.0, 1.5] \\
4--5 (23) & 38.3 [22.6, 53.9] & 5.2 [0.9, 10.4] & 0.9 [0.0, 2.6] & 0.0 [0.0, 0.0] & 0.0 [0.0, 0.0] & 0.0 [0.0, 0.0] \\
$\geq$6 (17) & 32.9 [16.5, 50.6] & 3.5 [0.0, 9.4] & 2.4 [0.0, 5.9] & 0.0 [0.0, 0.0] & 0.0 [0.0, 0.0] & 0.0 [0.0, 0.0] \\
\bottomrule
\end{tabular}
\endgroup

\end{table}
Appendix Figure~\ref{fig:failure-primary}(c) shows the lower triangle of the
Spearman property matrix for active tasks; the omitted upper triangle is
symmetric. Demand, region count, rules, vertices, and recorded first-planner
prompt tokens co-vary. Coordinates denotes visible coordinate pairs; D is a
rule composite. These associations do not isolate causal task effects.

To examine associations conditional on other task properties, we fit
exploratory L2-penalized logistic regressions
($C=10$). Region and vertex-count VIFs are about
7.70 and 6.28. The fit substitutes zero for one missing transistor count. Co-varying
predictors and this imputation make the coefficients descriptive associations.
In these regressions, some Kimi geometry/architecture intervals exclude zero.

\FloatBarrier
\section{Construction methods and provenance units}
\label{app:construction}
\paragraph{Composition groups.}\label{app:composition-groups}
Table~\ref{tab:composition} uses disjoint active-manifest groups, each task
counted once. Top-gated base (55) is the base inventory, including four
controls. Hole and architecture probes (19) comprise 15 hole tasks and four
architecture probes. Gate/contact variants (18) include ten local back-gated,
four edge-contacted, and four dual-gated tasks. Repeated devices and circuits
(18) add repeated-device and circuit templates. Three material roles (12)
use three proxy material-role labels. Scene/constraint variants (6) adapt
existing templates to CMOS-pair, dense multi-flake, or tight-rule scenes.
These groups describe construction provenance. Region counts and
transistor-demand medians are calculated per task. The unspecified demand belongs
to L2/case30 in probes. Percentages may sum to 100.1\% after rounding.

\paragraph{Extraction and selection.}
MaskTerial semantic-mask connected components are used as flake identities,
merging per-thickness fragments. Masks are $1920\times1200$ pixels; components
touching the image border, with maximum extent below 60 pixels, or with fewer
than 300 pixels are removed. The nominal scale is $0.293\,\mu$m/pixel, without
independent per-image calibration; upstream descriptions suggest approximately
$0.39$ or $0.5\,\mu$m/pixel, so reported absolute dimensions are nominal.
KLayout smoothing occurs first at 1.5 pixels
($0.44\,\mu$m), then at 1.0 pixel ($0.293\,\mu$m) during task generation;
hole rings use $0.15\,\mu$m. Measuring fidelity after both passes requires a separate comparison beyond
the single-pass library sweep. The active contours have a median of 13 vertices.
The main contour pool contains 3,220 one-component, hole-free graphene shapes,
no larger than $800\,\mu$m$^2$ and with at least four vertices. Smoothed
solidity (polygon area divided by convex-hull area) defines shape tiers G1
($\geq0.90$), G2 (0.75--0.90), and G3 ($<0.75$).
Hole probes are selected separately. Shape tiers describe contour solidity.

Figure~\ref{fig:interface}(a) plots all 3,220 eligible records, including duplicates; there are
2,679 distinct stored source polygon arrays. Solidity is simplified polygon
area divided by convex-hull area, rounded to four decimals. Fixed strata
G3 ($<0.75$), G2 (0.75--0.90), and G1 ($\geq0.90$) contain
644, 1,313, and 1,263 records. These counts give the observed frequencies in the contour pool. The nominal G1/G2/G3 schedule is
30/40/30 within each template level; non-top-gated templates force G1 to restrict
shape tier in those architecture comparisons. Within-tier shuffled queues are
drawn without replacement until replenished; rejected capacity candidates
are consumed. Hole probes follow a separate path: 65 holed records, 50 with
area 60--800 $\mu$m$^2$, 26 eligible records, and 15 stored tasks.
\paragraph{Spatial fitting.}
For a Thomas process, Poisson parents have intensity $\kappa$, offspring
counts have mean $\mu$, and Gaussian offsets have scale $\sigma$. Its intensity
is $\lambda=\kappa\mu$ and
\begin{equation}
 K(r)=\pi r^2+\frac{1-\exp[-r^2/(4\sigma^2)]}{\kappa}.
\end{equation}
The stored fit pools images with at least two retained flakes and uses
border-corrected empirical $K$. It minimizes squared discrepancies between
fourth roots of empirical and model $K$ over radii 20--420 pixels, with 16
Nelder--Mead starts, then sets $\widehat\mu=\widehat\lambda/\widehat\kappa$.
Conditioning on multi-flake fields affects both intensity and clustering.

\begin{table}[h]\centering\small
\caption{Stored Thomas-process fit parameters; weights are proportional to
centroid counts. Image counts refer to the records used for fitting.}
\label{tab:thomas}
\begin{tabular}{lrrrrrr}\toprule
Dataset & Images & Centroids & $\kappa$ ($10^{-5}\,\mu$m$^{-2}$) & $\sigma$ ($\mu$m) & $\mu$ & Weight\\\midrule
GrapheneH & 259 & 783 & 1.375 & 12.07 & 1.112 & .365\\
GrapheneL & 314 & 855 & 1.605 & 14.77 & .858 & .399\\
GrapheneM & 166 & 505 & 1.162 & 12.14 & 1.324 & .236\\\bottomrule
\end{tabular}\end{table}

\paragraph{Placement and screening.}
Multi-region scenes select a fitted component and a window with area
$n/\widehat\lambda$ and aspect ratio 1.6. Parents are simulated on an expanded
window and offspring outside the scene are dropped. The sampler retries until
at least $n$ points exist, then selects exactly $n$, applies rotations, and
requires separation and in-window bounds. The resulting distribution reflects the fitted Thomas process, point-count
selection, and placement screening. Polygon dilation
by $0.25\,\mu$m enforces a $0.5\,\mu$m gap; the final scene is translated to a
$5\,\mu$m margin. No scale or mirror transformation is applied in this main
placement loop. Single-region
tasks bypass spatial sampling but receive a uniform random rotation. The contour
and final orientation must admit reference construction under the hidden demand
and implemented rules: capacity is the largest transistor count, up to eight,
that the reference constructor's transistor-distribution routine can place on
the contour's medial-axis spine under the task rules, evaluated in library
orientation at draw time and again in situ after placement. Retry limits are
fixed in the generator: parents are simulated on the window padded by
$4\sigma$; up to 500 centroid draws reach $n$ points; on a bounds or
separation conflict a contour is re-proposed around its own cluster parent, up
to 25 times per pattern and up to 20 patterns; a capacity-qualified
contour is sought in up to 150 tier draws; and placement plus screening is
repeated up to 40 times, swapping out a failing contour every eighth attempt.
Exact sampling replay and acceptance-rate estimates would require rejection
counts, latent parents, and per-task placement seeds beyond the stored records.
The seed-0 panel illustrates the placement procedure.

Thirteen active multi-region tasks underwent a later tight-window transform:
ten at 0.30, two at 0.35, and one at 0.55. This historical branch is distinct
from the main placement loop; L3/case8 is a stored tightened example.
\paragraph{Counting-unit reconciliation.}
The provenance export for the 128 tasks lists 317 ``source images'' that are
dataset-prefixed keys, not bare filenames. The completed join of task flake IDs to library
records gives 370 placements, 352 dataset-prefixed flake IDs, 317 distinct
(dataset, split, filename) keys, and 290 distinct filename basenames. Repeated
names across shards must not silently be counted as new physical sources or
deduplicated as identical physical sources without evidence. A filename/component
join produces 343 keys, which likewise is not an independently established
physical-flake count. We did not perform source-image hash deduplication.
The join and reconciliation files are retained with the manuscript's source
mapping. All 170 MoSe$_2$/WSe$_2$-labeled placements carry graphene identities.
The three MaskTerial graphene datasets share micrographs: 365/409 GrapheneM
images with retained flakes reuse GrapheneH filenames, and 558/748 GrapheneM
contour records are exact GrapheneH twins. The 352 active dataset-prefixed
flake identities correspond to 347 distinct stored contours. Thus the
dataset-prefixed image groupings in Appendix~\ref{app:statistics} understate
dependence; basename-based groupings are a sensitivity analysis, not verified
physical-specimen identities.

\paragraph{Figure provenance.}
Figure~\ref{fig:interface}(b) illustrates a separate sampling draw; panel (d)
shows the stored task scene at its original coordinates.
Its contacts and gates come from the stored L3/case8 reference GDS, rechecked
against the current canonical verifier and fixed-input round-trip XOR.
Panel (b) shows four retained offspring in two local parent neighborhoods
from a six-flake seed-0 draw in an $835\times522\,\mu$m window. The neighborhoods
use separate views at a common display scale. The seed was
selected as the first successful draw with a parent retaining at least two
offspring. This selection illustrates a multi-offspring neighborhood. Open
circles show the selected pre-jitter proposals. Solid dots are final positions,
and dotted arrows join moved proposals to their retained positions.
The separation threshold is $0.5\,\mu$m. Screening requires
reference capacity to meet hidden demand and final-orientation feasibility
agreement, with redraws on failure. The illustrative draw demonstrates placement; task-specific reference
screening is performed during benchmark construction.
For context, the fitted GrapheneL component often produces singleton clusters:
$P(N=1\mid N>0)=\mu/(e^\mu-1)=0.632$ at $\mu=0.858$, before window,
count-selection, and placement filtering. L3/case8 requires two transistors
per region. Panel (d) uses exact stored positions
and reference layers; the device crop $[8.5,16.5]\times[34,43.5]\,\mu$m
and the overview each have a nominal scale bar. Contact shapes are shown
without source/drain assignments.
\FloatBarrier
\section{Effective verifier and preservation audit}
\label{app:verifier}
The implementation labels the structural checks LVS; we call them structural
checks to distinguish them from schematic-driven electrical LVS with device and
net extraction. The database unit is $0.001\,\mu$m. Coordinate conversion has truncation/rounding
asymmetry across realization paths; DRC deficit comparisons use approximately
$1.001$ nm tolerance. Success is evaluated on this grid with the stated numerical tolerances. Required layers follow design order
with reference entries excluded. Pad-cover layers on 32 tasks and dielectric
layers on two are checked for nonempty geometry. Their electrical function
requires separate validation.

Plain transistor counts are lower bounds derived from gate--material crossings
and contact overlap; surplus crossings are allowed. A wrap-around contact may
satisfy both sides without independently extracted source/drain nets. Circuit
checks instead impose exact load-derived counts, per-flake instances, and
series/parallel contact-sharing templates. These checks assess geometric device patterns; electrical netlist validation
requires a separate evaluation. Injected via-isolation checks are part of the effective
contract even though they are absent from the raw task JSON. Fixed-input equality is a separate final artifact condition.
Expected fixed regions are initialized from task polygons and holes, then
GDS-round-tripped before XOR comparison with the unchanged submitted layout.
This includes serialization seams without silently repairing altered inputs.

The completed six-model fixed-input audit finds 38 Qwen violations (32 for
235B, six for 32B). One otherwise passing artifact is invalidated; the other
37 already fail geometric checks. All reported endpoints include preservation.

An isolated copy of the benchmark verifier re-scores all 128 reference GDS files
and all 7,680 current-panel attempt artifacts with agreement on every recorded
verdict; all references pass and preserve fixed inputs. Isolated release checks
also report 15 passing predicate tests, one expected
limitation, and 12 positive/negative scoring cases. These checks validate the listed predicates and scoring cases.

\FloatBarrier
\section{V1--V2 construction lineage and comparison boundary}
\label{app:version-design}
\begin{table}[H]
\centering\small
\caption{Task-design lineage. Current stored definitions are compared; historical
scores do not form a controlled v1/v2 experiment.}
\label{tab:version-design-audit}
\begin{tabular}{@{}p{0.20\linewidth}p{0.34\linewidth}p{0.36\linewidth}@{}}
\toprule
 & V1 & V2 \\
\midrule
Task set & 130 stored tasks & 128 active of 145 stored \\
Source geometry & Procedural polygon families & Microscopy-derived graphene contours; proxy material roles \\
Lineage & Original templates & 113 descendants and 15 added hole probes \\
Regions/task & Median 2; range 1--10 & Median 2; range 1--10 \\
Device demand & Median 4; range 1--24; two unspecified & Median 3; range 1--24; one unspecified \\
Hole tasks & 0 & 15 \\
Circuit tasks & 31 & 26 \\
Shared interface & \multicolumn{2}{p{0.70\linewidth}@{}}{Coordinates, visible DRC, hidden verification criteria, structured geometry actions and deterministic checks} \\
Success contract & Historical scorer and report-specific infeasibility labels & V2 checks plus saved fixed-input preservation \\
\bottomrule
\end{tabular}
\par\smallskip\raggedright
Demand is summed over specified regions before task-level aggregation; missing
values are excluded. Plain transistor counts are lower bounds in both retained
scorers; circuit templates have separate exact count/topology checks. Reused
filenames do not denote unchanged tasks or paired historical trials.
\end{table}

The retained V1 inventory has 130 procedural tasks. All 130 identities have
V2 descendants, but only 113 enter the 128 scored tasks; 15 newly added hole probes
complete the active cohort. Every inherited material-region geometry changes.
Some descendants also change scene counts, material roles, or hidden device
demand. The coordinate interface, structured
geometry output, verifier checks, C0--C2 scaffolds, workflow implementations,
and prompts predate V2, as do the DRC rule sets of all 113 inherited tasks and
the circuit conventions and scored counts of all 26 circuit tasks. V1 reported
five attempts per task for one condition only. V2 contributes the
microscopy-derived contours, hole probes, selected scene conversions, cohort
curation, five-attempt evaluation of every reported arm with the all-five
endpoint, the fixed-input preservation check, and the full reference and attempt
re-score. The inherited scoring rules retain the scope described above.

The retained definitions and conversion scripts establish design lineage, not
unchanged task pairs or exact submitted-era artifact identity. Historical model
runs differ in task inclusion, geometry, serving/reasoning, prompts and budgets,
harness/scorer revisions, and repeat counts; generation-time pins are incomplete.
We report V2 performance on its own task set. The 74-task residual-distractor
set is a separate geometry variant. A paired V1--V2 comparison would require
a common evaluation protocol and would measure the combined effect of the
geometric properties changed by contour replacement.

\FloatBarrier
\section{Historical and excluded comparisons}
\label{app:historical}
\begin{table}[h]\centering\small
\caption{Historical GLM-5 workflows, 640 attempts per arm. These reasoning-off
Bedrock Mantle configurations are separate from the current GPT5.6-Luna study and the
primary GLM-5 configuration with high reasoning.}
\label{tab:historical}\begin{tabular}{lrrrrr}
\toprule
Historical arm & Successes & pass@1 & pass@5 & M tokens & Agent runs \\
\midrule
Plan-and-Execute & 38 & 5.9 & 18.8 & 2.76 & 640 \\
ReAct-4 & 62 & 9.7 & 21.1 & 68.31 & 8279 \\
Reflexion-4 & 68 & 10.6 & 22.7 & 83.88 & 9854 \\
ExpeL-4 & 71 & 11.1 & 24.2 & 86.46 & 9636 \\
\bottomrule
\end{tabular}
\end{table}

The rescored historical ReAct-4 row has 62 successes and an older provider
output cap. An earlier ReAct-4 run with 66 in-loop successes is not used because
it lacks the same saved-layout audit. Two historical layouts alter fixed inputs;
both also fail geometric checks. Relative to historical PaE, the layer-by-layer
configurations use about 24.8--31.3 times the recorded tokens, with serving and revision
differences. Historical GLM-5 and current GPT5.6-Luna comparisons describe
their respective configurations and evaluation settings.

The wider historical matrix records 24 model/workflow configurations; $24\times640$
is the number of scheduled attempts; executed-attempt counts come from the
canonical analysis export.
The canonical analysis export contains 12,090 observations: 7,680 current-panel
attempts, 2,560 historical GLM attempts, and 1,850 attempts on the 74-task residual-distractor set.
Residual-distractor tasks change the fixed geometry and use separate campaigns
and routes, including local models and a mixed-route Kimi condition; they do
not enter current-panel rankings. Incomplete campaigns are nonrandom coverage
snapshots and are not added to complete-arm comparisons. Eight archived and
nine analysis-only task identities likewise remain outside the 128 scored
tasks; the historical ``holdout'' label denotes membership in that inventory.

\FloatBarrier
\section{Presentation provenance and release status}
\label{app:release}
Local plotting scripts produce PDF, SVG, and PNG assets and record
source hashes. The existing packages, task files, verifier, harness, campaign
configurations, and historical outcomes are preserved. V2 is the historical
artifact selected for release. The dataset will be released after publication.

The evaluator package for the reported V2 benchmark contains
128 unchanged task files, the scoring modules used for these results, an explicit
fixed-input check, eight GDS fixtures, an active-only membership checker,
pinned evaluator dependencies and checksums. The original all145 inventory is
preserved as provenance; eight archived and nine analysis-only files are absent.
The wrapper reports rule success, fixed-input preservation and overall success
separately. It excludes obsolete scorers, unknown-rights templates, exemplars,
raw provider responses, confidential reviews and campaign state.

The unpacked package was installed in a fresh Python 3.10.12/Linux
environment using its 23-package lock. All eight fixtures passed their expected
verdict assertions: four passing reference layouts, two recorded model outputs
(one success and one failure), a missing-required-layer failure, and a modified-input
control that passes rules but fails overall. These tests used no API credentials;
the smoke test denied Python socket connections, not kernel-level networking.
Installation required package downloads. This test validates installation and scoring for the listed fixtures.
The separate all-task re-score is described in Appendix~\ref{app:verifier}. 

\FloatBarrier
\section{Failure symptoms}
\label{app:failures}
\begin{figure}[t]\centering
\includegraphics[width=.97\textwidth]{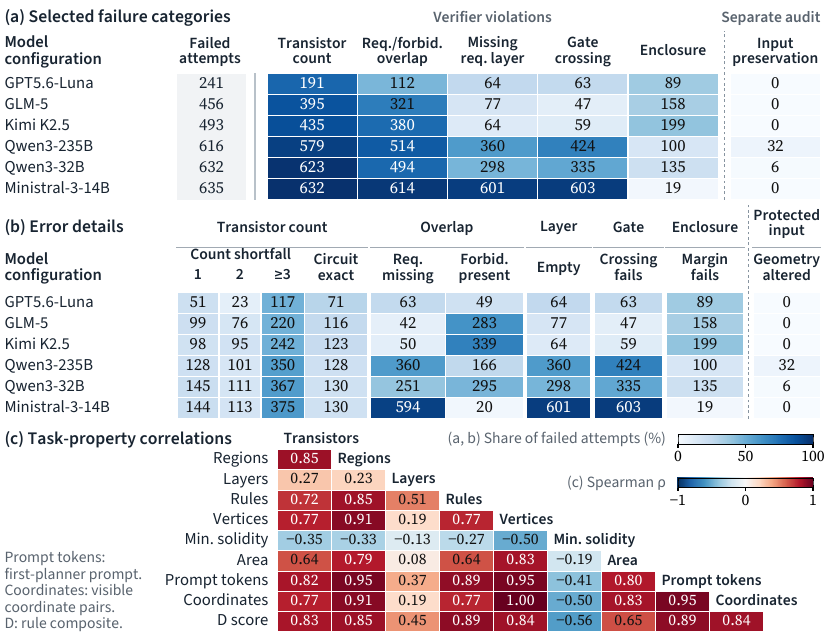}
\caption{Failure symptoms in the six ReAct-3/C0 configurations (640 attempts
each). (a,b) Counts of affected failed attempts, colored by the share of each
configuration's failed attempts; predicates overlap, and a low enclosure count
can reflect a check skipped when required geometry is absent. (c) Descriptive
Spearman correlations among task properties, not failure rates. Definitions
and applicability counts are in Appendix~\ref{app:failure-mapping}.}
\label{fig:failure-primary}
\end{figure}

\paragraph{Failure categories and counting units.}
\label{app:failure-mapping}
Figure~\ref{fig:failure-primary}(a) reports the broad counts. Affected attempts are deduplicated by model
arm, task identity, and attempt index within each category. Transistor count
maps to \texttt{transistor\_count}; required/forbidden overlap is the union of
\texttt{must\_overlap} and \texttt{must\_not\_overlap}; missing required layer,
gate crossing, and enclosure map to \texttt{layer\_not\_empty},
\texttt{must\_cross}, and \texttt{min\_enclosure}. Fixed-input integrity comes
from the separate saved-layout preservation audit, not a verifier predicate.
It remains part of success even when geometry checks also fail.
The complete saved-message taxonomy classifies every attempt, but 26 failed
attempts have only predicates omitted from Figure~\ref{fig:failure-primary}.
Panel (b) separates ordinary transistor-count shortfalls from circuit
exact-count and overlap predicates. For enclosure, 326/345 applicable
Ministral layouts meet the explicit empty-layer skip condition; a low
violation count can therefore reflect incomplete geometry rather than a
passing check.

Panel (b) uses ordinary \texttt{[LVS]} expected/found messages to calculate
a count shortfall: the sum of positive expected-minus-found deficits over
distinct regions. Repeated identical messages for one region are counted once;
region names and expected counts are checked against the canonical task. The
1, 2, and $\geq3$ bins are mutually exclusive and recover the ordinary-count
parent. They measure missing valid transistor counts under the implemented
checks, not absent shapes or a demand-adjusted measure of error severity.
\texttt{[Circuit]} exact-count mismatches remain a separate overlapping column.
Ordinary surplus counts are not errors. Recorded expected/found values identify two above-exact-count
attempts (one GLM-5, one Kimi K2.5); each also has a below-exact violation in
another region. Circuit-below/circuit-above attempt counts are respectively
71/0, 116/1, 123/1, 128/0, 130/0, and 130/0 in model-row order. All circuit-count
violations here co-occur with an ordinary minimum-count violation. Required
versus forbidden overlap uses the two distinct canonical predicates directly;
their counts (Table~\ref{tab:systematic-applicability}) can sum to more than
the parent union. The other four columns repeat the
recorded missing-layer, crossing, enclosure, and input-preservation categories
without claiming further subdivisions. In the compact headers, ``Empty'' means
a required layer has no geometry; ``Crossing fails'' means a required crossing
is absent or insufficient; ``Margin fails'' means insufficient enclosure; and
``Geometry altered'' means the protected input geometry was altered.

The failure analysis checks all 3,840 canonical records. For GPT5.6-Luna it uses
saved-layout audit messages; for the five other models it joins recorded final
error summaries to the canonical audit's source-record paths. All verifier
predicate multisets agree, including repeated instances. Twelve separately
recorded execution-error messages are identified and excluded only from this
predicate comparison; their attempts remain in the cohort. Count types and
expected/found values parse for every relevant message, with zero unclassified
attempts. This verifies recorded-message concordance, not a new layout rescore
or complete reproduction of generation-time prompts. Both panels use all
failed attempts as denominator, not only attempts subject to a particular rule.
These are overlapping observed symptoms, not causal categories.

\subsection{Systematic error and revision analysis}
\label{app:systematic-errors}
This retrospective analysis uses the same 3,840 completed ReAct-3/C0 attempts.
All 128 task hashes and saved-layout hashes agree with the canonical records;
all 3,840 saved logs are readable and their final predicate multisets agree.
The logs distinguish post-planning Critic checks, post-restore checks, final
verification, and infeasibility declarations. Restore feedback is not a new
planner round. Fixed-input preservation is a separate final audit, not an
intermediate Critic predicate. Twelve recorded execution/validation diagnostics
remain separate from final geometric outcomes; no attempt is removed for them.
All 3,073 failed attempts have a recorded final predicate or integrity violation.
Table~\ref{tab:systematic-omitted} lists predicates omitted from the main figure.
This analysis uses the recorded outcomes.

\paragraph{Taxonomy and counting units.}
The minimum-count check also runs on circuit tasks: ``ordinary'' describes
\texttt{[LVS]} lower-bound semantics, not a non-circuit subset. Independent
\texttt{[Circuit]} messages enforce exact template counts. The three shortfall bins partition minimum-count failures, whereas circuit
mismatches can overlap them. The machine-readable taxonomy retains every generated predicate, source field,
code location, applicability, counting unit, and deduplication rule.

\begin{table}[htbp]\centering\small
\caption{Additional final predicates omitted from Figure~\ref{fig:failure-primary}.
Cells count affected attempts, deduplicated within predicate. ``Only omitted''
counts failed attempts with none of the six displayed categories; categories
elsewhere in the table can overlap. Boundary means required boundary contact.}
\label{tab:systematic-omitted}\begin{tabular}{lrrrrrr}
\toprule
Model & Separation & Spacing & Width & Boundary & Topology & \shortstack{Only\\omitted} \\
\midrule
GPT5.6-Luna & 62 & 18 & 12 & 1 & 0 & 3 \\
GLM-5 & 143 & 8 & 5 & 3 & 0 & 6 \\
Kimi K2.5 & 229 & 28 & 19 & 2 & 1 & 6 \\
Qwen3-235B & 165 & 52 & 22 & 9 & 0 & 7 \\
Qwen3-32B & 231 & 74 & 36 & 11 & 0 & 4 \\
Ministral-3-14B & 6 & 0 & 7 & 21 & 0 & 0 \\
\bottomrule
\end{tabular}

\end{table}

\paragraph{Prevalence, applicability and skipped checks.}
Companion tables give affected attempts, affected tasks, fractions of all
640 attempts and of failed attempts, and task-defined applicability.
Effective rules include injected via/gate isolation pairs. Applicability is
not the same as an evaluated check. We read layer occupancy from the saved
GDS files without running the verifier: enclosure, inter-layer separation,
and forbidden-overlap checks explicitly skip empty participating layers.
Width/spacing functions operate on empty regions without an explicit skip,
giving vacuous checks. Successful execution of every individual rule is not
instrumented; branch eligibility inferred from geometry is labeled separately.
Circuit exact-count messages identify topology branches skipped by the
checker; execution of remaining topology branches is not established by an
absent error alone. Unknown execution status is retained in the export.

Enclosure is specified on 69 tasks, giving 345 applicable attempts per model.
Ministral-3-14B has 19 enclosure errors among 635 failures, but 326/345
applicable final layouts meet the empty-layer skip condition; all remaining
19 have recorded enclosure errors. Thus low conditional frequency does not
establish better enclosure reasoning. Similar percentages among failed
attempts can also represent very different absolute counts and failure totals.

\begin{table}[htbp]\centering\small
\caption{Overlap intersections and enclosure applicability. Required/forbidden
columns count missing required/present forbidden overlaps; ``Both'' is their
attempt-level intersection. Their union recovers Figure~\ref{fig:failure-primary}.
Enclosure counts violations; ``Skipped / 345'' counts applicable attempts whose
saved geometry satisfies the explicit empty-layer skip guard, not passed checks.}
\label{tab:systematic-applicability}\begin{tabular}{lrrrrr}
\toprule
Model configuration & Required & Forbidden & Both & Enclosure & Skipped / 345 \\
\midrule
GPT5.6-Luna & 63 & 49 & 0 & 89 & 48 \\
GLM-5 & 42 & 283 & 4 & 158 & 35 \\
Kimi K2.5 & 50 & 339 & 9 & 199 & 36 \\
Qwen3-235B & 360 & 166 & 12 & 100 & 242 \\
Qwen3-32B & 251 & 295 & 52 & 135 & 209 \\
Ministral-3-14B & 594 & 20 & 0 & 19 & 326 \\
\bottomrule
\end{tabular}

\end{table}

\paragraph{Co-occurrence and incomplete construction.}
Pairwise intersections, unions, conditional rates and Jaccard ratios are
computed on deduplicated failed-attempt sets. Exports also give the number of
categories per failure and every observed combination; the findings report
lists frequent combinations with an explicit remainder. These are descriptive
associations, not causal dependence. Missing required layers overlap count,
crossing and required-overlap failures (Table~\ref{tab:systematic-missing}).
The checks help explain these overlaps: an empty intersection fails
required-overlap and crossing checks, and no qualifying gate/contact crossing
can contribute a valid transistor. Missing an arbitrary required layer does
not necessarily imply these failures; missing pad-cover geometry, for example,
is distinct from missing contacts. The observed overlaps do not prove a
temporal causal sequence.

\begin{table}[htbp]\centering\small
\caption{Missing-layer failures and their intersections with other violations.
Each intersection uses the row's ``Missing'' count, not 640, as denominator;
Required/Forbidden denote overlap checks.}
\label{tab:systematic-missing}\begin{tabular}{lrrrrr}
\toprule
Model configuration & Missing & $\cap$ Count & $\cap$ Crossing & $\cap$ Required & $\cap$ Forbidden \\
\midrule
GPT5.6-Luna & 64 & 64 & 63 & 63 & 0 \\
GLM-5 & 77 & 75 & 36 & 37 & 23 \\
Kimi K2.5 & 64 & 61 & 35 & 36 & 22 \\
Qwen3-235B & 360 & 354 & 324 & 324 & 35 \\
Qwen3-32B & 298 & 297 & 211 & 196 & 99 \\
Ministral-3-14B & 601 & 599 & 588 & 591 & 6 \\
\bottomrule
\end{tabular}

\end{table}

\paragraph{Task-property breakdowns.}
The same outcome-independent partitions are used for every model: demand
1/2--4/5--9/10--24 has 29/52/26/20 tasks, with one unspecified task retained
as missing; 1--2/3/4--5 required layers has 57/36/35; 1/2/3--4/5+ target
regions has 49/25/29/25; circuit/non-circuit has 26/102; and holes/no holes has
15/113. Complete
model-by-category subgroup counts and denominators are saved in
\texttt{subgroup\_prevalence.csv}; a compact count-failure matrix appears in the
findings report. These properties overlap and small strata can be unstable.
Missing demand is not zero.
These summaries are retrospective and descriptive; no significance tests
or additional uncertainty intervals are introduced.

\paragraph{Observed revision trajectories.}
All logs match unique attempts and consistent Critic-check counts. A declaration
or execution error can precede the first check: 62/11/20/75/43/483 attempts
have no intermediate check in model-row order, but all have final verification.
Table~\ref{tab:systematic-trajectories} partitions 640 attempts per model by
final outcome and observed termination.
GPT5.6-Luna has 223 final successes after the first check and 176 after revision,
giving 399 successes in total. These are observed workflow outcomes, not
an experiment isolating the benefit of feedback. One Qwen3-235B layout passes
its first Critic check but fails the final protected-input audit; it belongs
to the integrity column rather than first-check final successes.

\begin{table}[htbp]\centering\small
\caption{Observed outcomes (640 per row). First/Revised count final successes
after first/later checks; other columns count final failures by round cap,
declaration, execution error, or integrity audit. Stops are not adjudicated causes.}
\label{tab:systematic-trajectories}\begin{tabular}{lrrrrrr}
\toprule
Model configuration & First & Revised & Cap & Declaration & Execution & Integrity \\
\midrule
GPT5.6-Luna & 223 & 176 & 178 & 63 & 0 & 0 \\
GLM-5 & 73 & 111 & 421 & 34 & 1 & 0 \\
Kimi K2.5 & 48 & 99 & 460 & 33 & 0 & 0 \\
Qwen3-235B & 3 & 21 & 365 & 250 & 0 & 1 \\
Qwen3-32B & 3 & 5 & 502 & 129 & 1 & 0 \\
Ministral-3-14B & 4 & 1 & 52 & 573 & 10 & 0 \\
\bottomrule
\end{tabular}

\end{table}

Later-round rates condition on reaching an observed check
(Table~\ref{tab:systematic-rounds}). Across 4,677 consecutive observed-check
transitions, 1,390 both remove and introduce at least one predicate type;
transitions are repeated observations within attempts, not independent samples.
Full-layout replanning can resolve some violations while introducing others. Predicate-specific exports
give previous-violation denominators for resolution/persistence, and
previous-absence denominators for newly appearing violations.

\begin{table}[htbp]\centering\small
\caption{Critic passes / observed checks by round; later-round denominators
are selected. ``Resolved + new'' counts both changes / adjacent check pairs.
Fixed-input integrity is a separate final audit.}
\label{tab:systematic-rounds}\begin{tabular}{lrrrr}
\toprule
Model configuration & Check 1 & Check 2 & Check 3 & Resolved + new \\
\midrule
GPT5.6-Luna & 223/578 & 126/354 & 50/228 & 121/582 \\
GLM-5 & 73/629 & 62/543 & 49/470 & 246/1013 \\
Kimi K2.5 & 48/620 & 61/567 & 38/498 & 222/1065 \\
Qwen3-235B & 4/565 & 7/463 & 14/379 & 359/842 \\
Qwen3-32B & 3/597 & 5/546 & 0/502 & 374/1048 \\
Ministral-3-14B & 4/157 & 1/75 & 0/52 & 68/127 \\
\bottomrule
\end{tabular}

\end{table}

Illustrations use the first match in fixed model/task/attempt/transition order.
GPT5.6-Luna L1/case1 attempt 2 resolves enclosure and succeeds; L2/case37
attempt 4 resolves separation/forbidden overlap but adds count failure;
L2/case14 attempt 1 retains count failure. Log paths, line numbers, candidate
counts, and the selection rule are in \texttt{representative\_traces.csv}.
These selected examples illustrate revision behavior; they do not estimate its prevalence.

\FloatBarrier
\section{Data card}
\label{app:datacard}
The card records provenance, intended use, limitations, and maintenance in line
with dataset-documentation principles \citep{gebru2021datasheets}.

\paragraph{Purpose and appropriate use.}

2D-FET-Bench evaluates construction of GDSII layouts around irregular 2D-material contours using textual requirements and polygon coordinates. It measures success under a particular geometric/topological verification implementation plus fixed-input integrity checks.

Appropriate uses include descriptive comparisons of layout-generation systems under explicitly matched task, prompt and scoring versions, and analysis of geometric construction failures. A pass establishes compliance with the implemented geometric and structural predicates and preservation of fixed inputs. Electrical function, netlist equivalence, and fabrication readiness require separate validation.

\paragraph{Composition and identifiers.}

The 128 active tasks fall into the six provenance groups of Table~\ref{tab:composition}: top-gated base (``core'', 55), hole and architecture probes (19), gate/contact variants (18), repeated devices and circuits (18), three material roles (12) and scene/constraint variants (6).

The full manifest contains 145 distinct task files: 128 active, eight archived and nine analysis-only identities under the historical holdout label. Four controls are already part of core. These labels record inventory membership; evaluation uses the full active set. Core plus probes defines a 74-identity subset; residual variants of those identities are different task contents and must be reported separately.

Canonical membership is in tasks/v2/manifest.json. The accompanying TASK\_INDEX.csv is a derived active-only index, not a replacement for that source manifest. Identify a task using pool, relative path and SHA-256, as well as its JSON id; clean/residual variants can reuse IDs. Do not select tasks by recursively loading all 145 files.

The active set contains 370 target regions; 170 use MoSe$_2$/WSe$_2$ proxy labels. Source images and templates recur. The 317 image IDs and 352 flake IDs identify dataset records. Shared images across source shards can give multiple records for one physical specimen.

\paragraph{Task interface and units.}

Each task is JSON with:

\begin{itemize}
\item id, name and description: identity and human-readable objective;
\item agent\_visible.environment: layer IDs/datatypes/status, material polygons, optional holes, architecture and design\_order;
\item agent\_visible.drc\_rules: geometric constraints supplied to the agent;
\item verification\_only: structural checks and feasibility metadata used by evaluation;
\item v2\_meta: source/construction metadata and rule-based labels.
\end{itemize}

The agent should receive the intended visible interface, not the complete verification\_only object. Read-only material layers are fixed inputs. In design\_order, @layer entries locate references rather than request construction. Layer numbers must be interpreted using each task, not a global layer-name assumption.

Coordinates are nominal benchmark micrometers, with a frozen export grid of 0.001 $\mu$m. Polygon conversion, path rounding and tolerance behavior belong to the numerical scoring contract. Preserving fixed inputs requires the documented GDS round-trip comparison, including hole serialization; arbitrary independent rounding is not guaranteed equivalent.

L denotes template/directory provenance. D is a rule-based composite involving task properties, including hidden demand, revised after model outcomes existed. Both labels describe task properties and construction provenance.

\paragraph{Provenance and construction.}

Target contours are derived from real graphene semantic masks in the MaskTerial source dataset (\url{https://doi.org/10.5281/zenodo.15765514}). MoSe$_2$ and WSe$_2$ labels assign device roles to the graphene-derived contours.

Construction extracts and smooths polygons, applies an assumed 0.293 $\mu$m/pixel conversion, and rotates/translates selected contours into synthetic scenes. There are two smoothing stages, approximately 1.5 pixels then 1 pixel; hole handling has additional processing. The same nominal scale is used across source images.

For multi-region scenes, a fitted Thomas process supplies placement proposals, followed by point-count selection/subsampling, separation constraints and reference-construction-based feasibility screening. The resulting distribution reflects both the fitted process and the screening steps. Forty-nine single-region tasks bypass point-process placement. The retained distribution represents the selected templates and reference-feasible layouts.

The manifest records generation seed 7, but that alone does not reproduce the final collection bit-for-bit after augmentation and historical changes. Source inputs, intermediate library, invocation history and per-task resampling information must accompany any exact-regeneration claim. Residual distractors belong to a separately identified variant pool; the graphene provenance statement above applies to target regions, not all residual geometry.

\paragraph{Scoring and evaluation scope.}

Success requires all applicable canonical verifier checks and unchanged fixed inputs. Required editable layers are checked for presence; via/gate isolation is added by the harness to the effective DRC contract. These derived rules are not all written in the task JSON. Use the canonical Critic plus the separate saved-GDS integrity audit.

Scope of the checks and task specifications:

\begin{enumerate}
\item Plain transistor\_count is a lower bound; exact circuit templates have different semantics. The checks are not full electrical LVS, and contact identity/shorts are not comprehensively modeled.
\item Pad-cover requirements in 32 active tasks and dielectric requirements in two can be presence-only. Presence is not functional layer construction.
\item Fixed-input equality is enforced by the final integrity audit, not by the Critic alone. Passing only a subset of checks is not benchmark success.
\item A model declaration of INFEASIBLE does not prove infeasibility. Workflow locking, filtering and feedback policies affect observed outcomes and must be reported with the harness.
\end{enumerate}

Reference layouts are built by code with access to the scorer criteria.

\paragraph{Evaluation and reproducibility.}

For five attempts per task, report pass@1 as mean attempt success, observed pass@5 as the fraction solved at least once, and all-five as the fraction solved in all five observed attempts. All-five measures consistency within the five observed attempts. Resample whole tasks jointly across arms for paired task-bootstrap analyses; trajectory numbers are not shared seeds. State dependence and multiplicity handling.

Report complete and incomplete campaigns separately, including task/scorer hashes, denominators, infrastructure outcomes and exclusions. Distinguish workflow attempts, completed agent.run calls and HTTP requests. Compare recorded tokens within each model; billing estimates require provider cost records.

Existing GDS evaluation does not inherently require an API key. The audits include credential-free scoring tests and a fresh Python 3.10 installation from the pinned requirements. These are bounded environment tests, not certification of a final downloadable archive or identical regeneration.

\paragraph{Biases, privacy and potential impact.}

Selection favors author-chosen architectures, graphene-derived shapes, selected contour sizes and layouts accepted by the reference construction process. Shared sources/templates, proxy material labels and hidden scoring demand limit transfer. Benchmark success may reflect discovering hidden requirements through feedback as well as constructing geometry.

The dataset contains material geometry and device specifications. Release review covers filenames, code, GDS metadata, and logs for identifying information, private paths, and credentials. Raw provider responses, model weights, environments and live-campaign state are outside the proposed minimal task release. Preserve required third-party attribution.

The benchmark supports reproducible research on layout construction and workflows that use verifier feedback. Reports should tie conclusions to the implemented checks, descriptive task labels, and information available during each workflow.

\paragraph{Rights, access and maintenance.}

The repository declares Apache-2.0 for code and CC BY 4.0 for task JSON files. Task contours derive from the MaskTerial releases, distributed under the MIT license \citep{maskterial_dataset,uslu2025maskterial}. One of these datasets reuses the images of the 2DMatGMM datasets, distributed under CC BY 4.0 \citep{uslu2023twodmatgmm}. The release will retain both notices and citations. Reference outputs, figures, and documentation will carry their own license statements.

The code and the full dataset will be released publicly.

Task-information or scoring changes require a new version, preserved old bytes and a migration note. Changed visible requirements require fresh evaluations for claims about the corrected version; existing historical results must remain labeled with their original contract.

\section{V2 task previews}
\label{app:task-gallery}
\begin{figure}[p]\centering
\includegraphics[width=\textwidth,height=0.88\textheight,keepaspectratio]{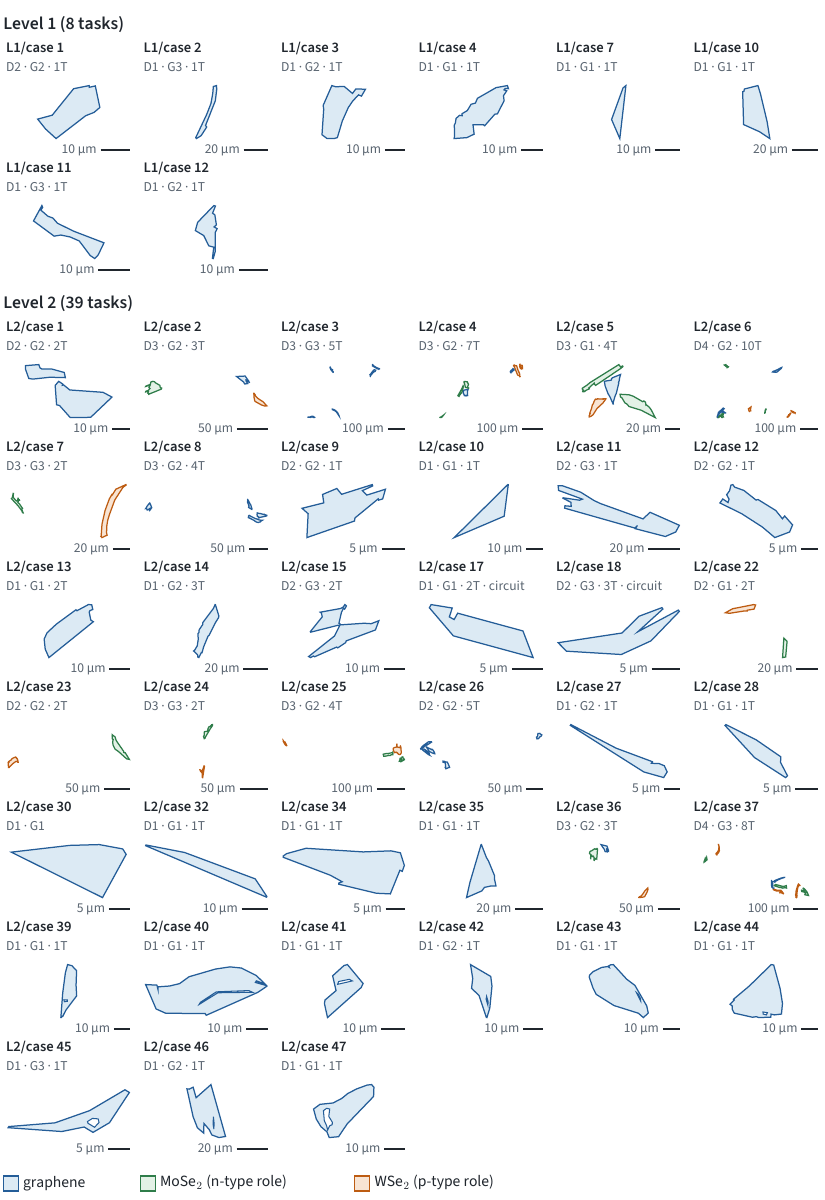}
\caption{V2 task previews for template levels L1 and L2 (47 tasks): input flake
contours colored by material label.}
\label{fig:task-gallery-1}
\end{figure}

\begin{figure}[p]\centering
\includegraphics[width=\textwidth,height=0.88\textheight,keepaspectratio]{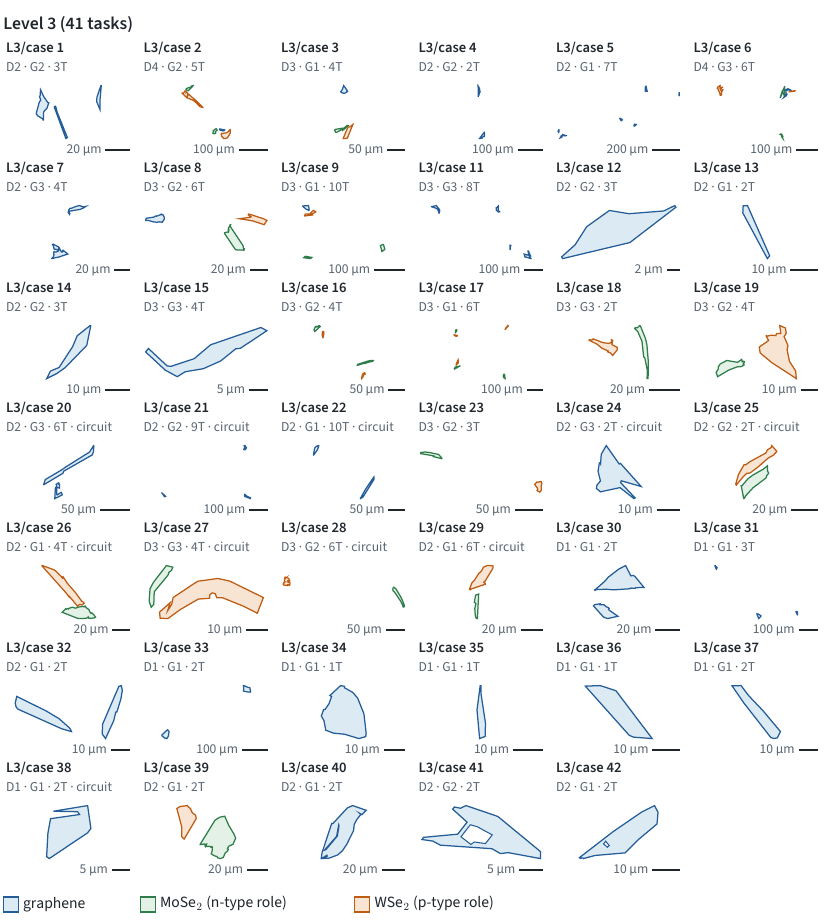}
\caption{V2 task previews for template level L3 (41 tasks): input flake
contours colored by material label.}
\label{fig:task-gallery-2}
\end{figure}

\begin{figure}[p]\centering
\includegraphics[width=\textwidth,height=0.88\textheight,keepaspectratio]{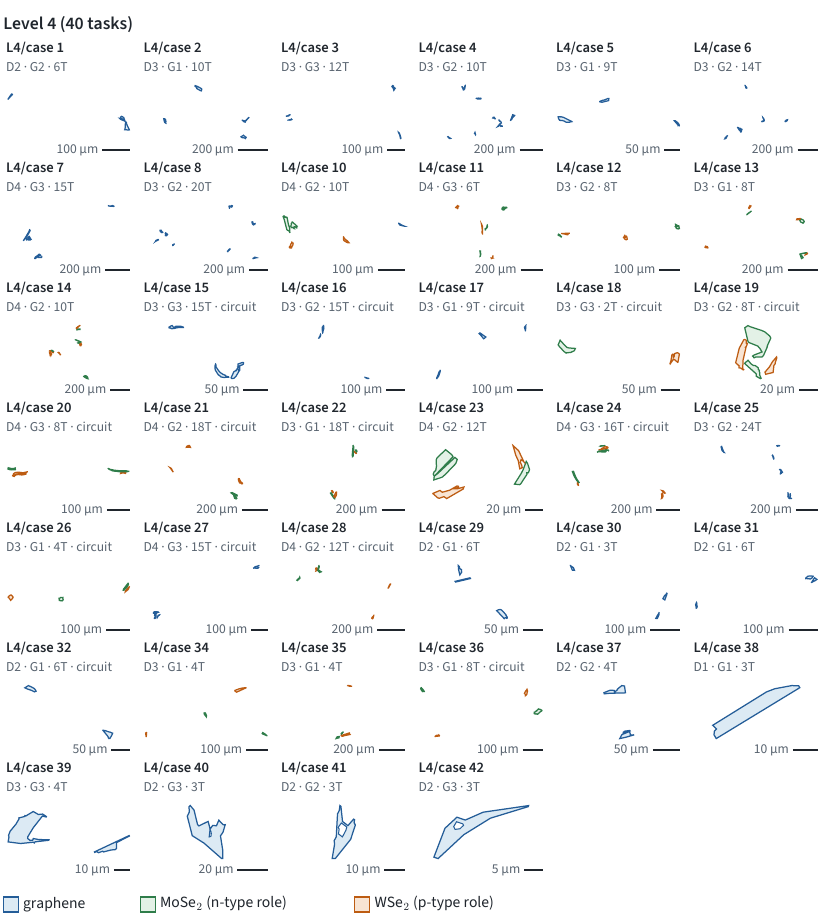}
\caption{V2 task previews for template level L4 (40 tasks): input flake
contours colored by material label.}
\label{fig:task-gallery-3}
\end{figure}

Figures~\ref{fig:task-gallery-1}--\ref{fig:task-gallery-3} show the input
flake contours of all 128 V2 tasks, grouped by template level. Each panel draws
the fixed flake polygons from the task file, colored by material label, with
interior holes in white. Editable layers are empty in the task input and are not
shown. Each panel has its own scale bar. The first label line gives the template
level and case number. The second gives the difficulty class (D), the shape tier
(G), the number of required transistors (T), and whether the task has a circuit
template. L2/case\_30 has no transistor-count criterion.
Figure~\ref{fig:interface}(d) shows a reference layout for L3/case\_8.
Figures are generated by \texttt{scripts/task\_gallery\_figure.py}.


\end{document}